\documentclass[sigconf]{acmart}

\renewcommand\footnotetextcopyrightpermission[1]{}
    
\copyrightyear{2021}
\acmYear{2021}
\setcopyright{acmcopyright}\acmConference[MM '21]{Proceedings of the 29th ACM
International Conference on Multimedia}{October 20--24, 2021}{Virtual Event, China}
\acmBooktitle{Proceedings of the 29th ACM International Conference on Multimedia
(MM '21), October 20--24, 2021, Virtual Event, China}
\acmPrice{15.00}
\acmDOI{10.1145/3474085.3475382}
\acmISBN{978-1-4503-8651-7/21/10}
    
\acmSubmissionID{1271}
    
\usepackage{amsmath}
\usepackage{bm}
\usepackage{multirow}
\usepackage{indentfirst} 
\usepackage{graphics}
\newcommand{\jcfi}{\textcolor[rgb]{0.93,0.51,0.93}}

\begin{document}

\title{Cross-Camera Feature Prediction for Intra-Camera Supervised Person Re-identification across Distant Scenes}


\author{Wenhang Ge$^{1,2}$, Chunyan Pan$^{1}$, Ancong Wu$^{1,2,*}$, Hongwei Zheng$^{4}$, Wei-Shi Zheng$^{1,2,3}$}

\makeatletter
\def\authornotetext#1{
\if@ACM@anonymous\else
    \g@addto@macro\@authornotes{
    \stepcounter{footnote}\footnotetext{#1}}
\fi}
\makeatother
\authornotetext{Corresponding author.}

\affiliation{
 \institution{\textsuperscript{\rm 1}School of Computer Science and Engineering, Sun Yat-sen University, Guangzhou, China}
  \institution{\textsuperscript{\rm 2}Pazhou Lab, Guangzhou, China}
 \institution{\textsuperscript{\rm 3}Key Laboratory of Machine Intelligence and Advanced Computing, Ministry of Education, China}
 \institution{\textsuperscript{\rm 4}Universtiy of Chinese Academy of Sciences, Xinjiang, China}
 \country{}
 }
\email{gewh@mail2.sysu.edu.cn; panchy8@mail2.sysu.edu.cn; wuanc@mail.sysu.edu.cn; hzheng@ms.xjb.ac.cn; wszheng@ieee.org}
\def\authors{Wenhang Ge, Chunyan Pan, Ancong Wu, Hongwei Zheng, Wei-Shi Zheng}
\renewcommand{\shortauthors}{Wenhang Ge, et al.}

\begin{abstract}
  Person re-identification (Re-ID) aims to match person images across non-overlapping camera views. The majority of Re-ID methods focus on small-scale surveillance systems in which each pedestrian is captured in different camera views of adjacent scenes. However, in large-scale surveillance systems that cover larger areas, it is required to track a pedestrian of interest across distant scenes (e.g., a criminal suspect escapes from one city to another). Since most pedestrians appear in limited local areas, it is difficult to collect training data with cross-camera pairs of the same person. In this work, we study intra-camera supervised person re-identification across distant scenes (ICS-DS Re-ID), which uses cross-camera unpaired data with intra-camera identity labels for training. It is challenging as cross-camera paired data plays a crucial role for learning camera-invariant features in most existing Re-ID methods. To learn camera-invariant representation from cross-camera unpaired training data, we propose a cross-camera feature prediction method to mine cross-camera self supervision information from camera-specific feature distribution by transforming fake cross-camera positive feature pairs and minimize the distances of the fake pairs. Furthermore, we automatically localize and extract local-level feature by a transformer. Joint learning of global-level and local-level features forms a global-local cross-camera feature prediction scheme for mining fine-grained cross-camera self supervision information. Finally, cross-camera self supervision and intra-camera supervision are aggregated in a framework. The experiments are conducted in the ICS-DS setting on Market-SCT, Duke-SCT and MSMT17-SCT datasets. The evaluation results demonstrate the superiority of our method, which gains significant improvements of 15.4 Rank-1 and 22.3 mAP on Market-SCT as compared to the second best method. Our code is available at \jcfi{\href{https://github.com/g3956/CCFP}{https://github.com/g3956/CCFP}}.

\end{abstract}






\maketitle
\vspace{-0.4cm}
\begin{figure}[ht]
    \includegraphics[width=8cm]{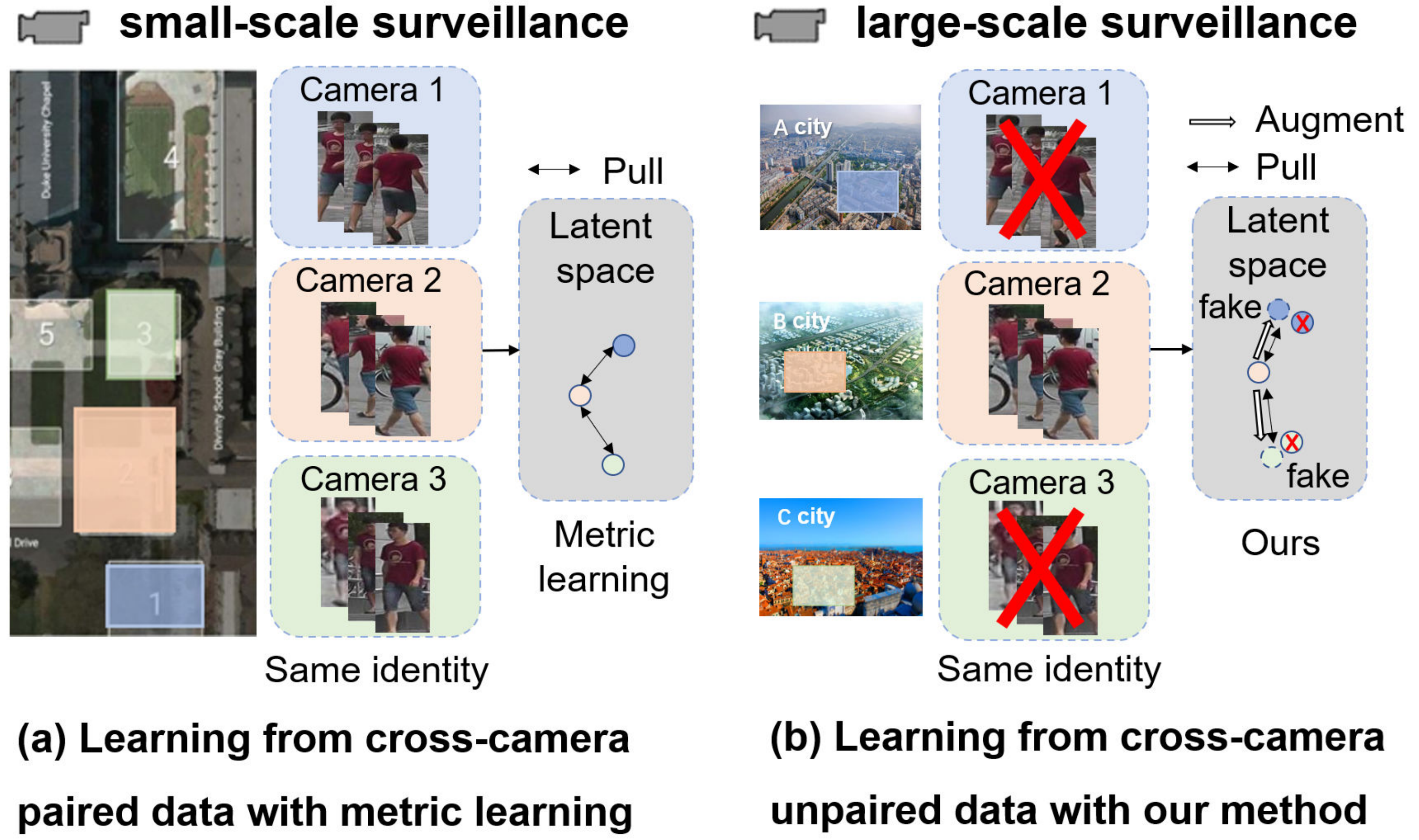}
    \vspace{-0.2cm}
    \caption{Difference between learning from cross-camera paired data and cross-camera unpaired data.
    (a) For Re-ID in adjacent scenes, cross-camera paired data can be captured for training.
    For most existing Re-ID methods, metric learning is performed on such data to learn camera-invariant representation.
    (b) For Re-ID in distant scenes, it is difficult to capture cross-camera paired data for training.
    To learn from cross-camera unpaired data in our method, we replace the missing real cross-camera features with transformed fake cross-camera features to mine self supervision information. % and minimize the distances to learn camera-invariant representation.
    }
    \label{skelent}
\end{figure}
%\vspace{-0.2cm}

\section{Introduction}
Person Re-Identification (Re-ID) has been widely investigated as a person retrieval issue across non-overlapping cameras.
Given a query person of interest, we aim to retrieve the person who shares the same identity with the query.
Due to its great value for social security, Re-ID in large-scale surveillance system is the development trend.
Thus, it is imperative to study  training method that requires fewer annotations for faster deployment to new scenes.

Most existing Re-ID methods focus on studying training method on
a small-scale surveillance system in which each pedestrian is captured in different camera views of adjacent scenes.
As shown in Figure \ref{skelent} (a), these methods perform metric learning on such cross-camera paired data, which is significant for overcoming cross-camera scene variations like viewing angle, background clutter, occlusion and body pose, so that the Re-ID model can learn how the same person appears differently in different cameras to extract camera-invariant representation.
When the Re-ID system is extended to a more large-scale one, pedestrian matching is not only required across adjacent scenes but also across distant scenes.
For example, a criminal suspect is escaping across different cities.
However, since most pedestrians appear in limited local areas, it is almost impossible for the same pedestrian to be captured in more than one of these distant scenes, so that it is difficult to obtain cross-camera paired training data, as shown in Figure \ref{skelent} (b).
%when tracking a pedestrian of interest across distant scenes, it is difficult to collect cross-camera paired data for training due to a limited local area of human activity.
%Since cross-camera scene variation will lead to poor generalization, training with existing cross-camera paired training set and then testing on the target domain is also infeasible.
Moreover, annotation for cross-camera pairs also requires massive costs.

One feasible way to reduce the annotation cost is labeling identities within each camera view, which is called intra-camera supervised person Re-ID \cite{tauld,zhu2019intra}.
However, most existing intra-camera supervised Re-ID methods need to associate underlying cross-camera positive pairs based on training data captured from adjacent scenes with overlapping identities. These methods are hardly applicable to distant scenes in large-scale Re-ID systems \cite{zhang2020single}.
%though there is no need to annotate cross-camera pairs, we still need to collect the training data including cross-camera pairs which is feasible under certain circumstances.
To overcome this limitation, we study intra-camera supervised person Re-ID across distant scenes (ICS-DS Re-ID), which utilizes cross-camera unpaired data with intra-camera identity labels for training.
This setting is also called Single Camera Training (SCT) \cite{zhang2020single}, which is still under-explored.
Lack of cross-camera paired data incurs a great challenge of how to establish the relation of unpaired samples among different cameras for learning camera-invariant representation.

%where each pedestrian will only be captured in one camera view with known tracklet as assumed in tracklet person Re-ID. We point out that the difference between SCT and tracklet person Re-ID\cite{tauld} is that there is no cross-camera tracklet association in SCT (i.e., cross-camera unpaired data). The core issue for this problem is how to guide the model to learn discriminative camera-invariant representation with cross-camera unpaired data. As many experiments show, deep metric learning loss which pulls the cross-camera positive pairs closer to the anchor and pushes the negative pairs away from the anchor in the embedding space endows the network an ability of extracting camera-invariant representation. It is the key for person Re-ID task. However, under our setting, this cross-camera cues are no longer exist. We are required to seek for other types of cross-camera information to provide guidance for learning camera-invariant representation. Zhang \emph{et al}\cite{zhang2020single} proposes a  triplet-analogy loss shown effective for SCT person Re-ID. However, cross-camera cues are still deficient. 

To address this issue, we propose a \textbf{C}ross-\textbf{C}amera \textbf{F}eature \textbf{P}re-diction (CCFP) self-supervised learning framework, which mines cross-camera self supervision information from camera-specific feature distribution to make connection between cross-camera unpaired data.
%by cross-camera feature augmentation to guide 
%by generating \textbf{C}ross-\textbf{C}amera \textbf{F}eature\textbf{s} (CCFs) via our proposed \textbf{C}ross-\textbf{C}amera \textbf{A}ugmentation module (CCA). The whole framework can be divided into three parts as follows.
To supplement missing cross-camera paired data in training set, we generate fake cross-camera positive pairs by cross-camera feature augmentation.
Then, we regard mapping a feature to its corresponding transformed cross-camera fake feature as cross-camera feature prediction, which can mine cross-camera self supervision information to guide camera-invariant feature learning.
Besides learning from global-level features, to further mine fine-grained cross-camera self supervision information, we introduce a transformer to automatically localize and extract local-level features. Joint learning of global-level and local-level features forms a global-local cross-camera feature prediction scheme.
Finally, we combine cross-camera self supervision and intra-camera supervision in our framework for learning camera-invariant representation. The main contributions of our work are summarized as follows:
\vspace{-0.05cm}
\begin{itemize}
    \item {We propose a cross-camera feature prediction method for mining cross-camera self supervision information to address the problem of lacking cross-camera paired data in the ICS-DS Re-ID setting.
    %To overcome the problem of lacking cross-camera cues on SCT person Re-ID, we propose a cross-camera predictor which is composed of $C$ camera-special batch normalization layers to do augmentation in the feature-level and obtain cross-camera features for global-level self-supervised learning.
    }
    \item {
    We further develop a global-local cross-camera feature prediction scheme for mining fine-grained cross-camera self supervision information.
    %Besides global-level self-supervised learning, to get rid of interference of background clutter, we selectively extract local feature by a transformer and aggregate it with global feature to form a global-local cross-camera feature prediction framework for mining fine-grained self supervision information to facilitate camera-invariant representation learning
    }
    \item {Extensive experiments on three benchmark datasets show that the proposed framework significantly outperformed the state-of-the-art methods in the ICS-DS Re-ID setting.
    }
\end{itemize}
\vspace{-0.2cm}

\section{Related Work}

\subsection{Fully Supervised Person Re-ID}
Fully supervised person Re-ID has made significant progress in recent years based on learning distance metric \cite{2012_CVPR_KISSME,Feature-Augmentation,xiong2014person,zheng2013RDC,wang2014person,wang2016person,Liu_2019_ICCV} , learning view-invariant discriminative feature \cite{gray2008viewpoint,2015_CVPR_LOMO,zheng2015scalable,bazzani2013symmetry} or
deep learning \cite{Li_DeepReID_2014b,subramaniam2016deep,2016_CVPR_JSTL,li2018harmonious,sun2018beyond,luo2019bag,chen2019abd, li2021combined}. However, fully supervised Re-ID requires a large amount of cross-camera identity annotations, which hinders fast deployment to new scenes. In this work, we investigate intra-camera supervised (ICS) Re-ID, where only intra-camera identity annotations are required, which are easily obtained with off-the-shelf person tracking algorithms \cite{luo2018trajectories}.

\vspace{-0.2cm}\subsection{Intra-Camera Supervised Person Re-ID}
In intra-camera supervised (ICS) Re-ID setting, identity labels are independently annotated within each camera and no cross-camera pair is labeled, which reduces the burden of annotation.

\vspace{0.1cm}\noindent \textbf{Intra-camera supervised Re-ID across adjacent scenes.}
Most existing ICS Re-ID methods assume that there are underlying overlapping identities across cameras in adjacent scenes in a small-scale surveillance system.
This is a problem of intra-camera supervised person re-identification across adjacent scenes (ICS-AS Re-ID).
Their main idea is that associating underlying cross-camera positive pairs. TAUDL \cite{tauld} separately treats each individual camera as a classification task and perform cross-camera positive pairs association.
MTML \cite{zhu2019intra} designs a cross-camera multi-label learning under a joint multi-task inference framework for discovering the cross-camera identity correspondence.
UGA \cite{graph_asso} proposes a graph association method to learn the underlying view-invariant representation. PCSL \cite{qi2020progressive} employs cross-camera soft labels for learning with weighted cross-entropy loss and triplet loss. ACAN \cite{qi2019adversarial} combines camera-alignment task and within-camera supervised learning task.
TSSL \cite{tssl} designs a comprehensive learning objective that combines tracklet frame
coherence, tracklet neighbourhood compactness, and tracklet cluster structure into a unified framework.
Wang \emph{et al.} \cite{wang2021towards} employ a non-parametric classifier for intra-camera and cross-camera learning.
%Existing intra-camera supervised person Re-ID assumes that cross-camera positive pairs are common which is appropriate under a small-scale surveillance systems, where each pedestrian is captured in different cameras of adjacent scenes (ICS-AS). However, in large-scale surveillance systems that cover larger areas, it is not easy to capture cross-camera positive pairs due to a limited range of human activities (i.e., cross-camera unpaired data).

\vspace{0.1cm}\noindent \textbf{Intra-camera supervised Re-ID across distant scenes.}
In this work, we study the problem of intra-camera supervised person person re-identification across distant scenes (ICS-DS Re-ID). Different from the assumption of ICS-AS that is valid for small-scale surveillance systems,
in large-scale surveillance systems, there is almost no cross-camera paired training data among distant scenes.
When learning from cross-camera unpaired data, the ICS-AS Re-ID methods based on underlying cross-camera positive pair association is not reliable, since they associate two cross-camera tracklets with different identities and thus such mismatch results in poor performance.
At present, the ICS-DS setting is only investigated in MCNL \cite{zhang2020single} and still under-explored.
MCNL \cite{zhang2020single} proposes a multi-camera negative loss, which pulls positive pairs closer to the anchor and pushes the within-camera negative pairs further than cross-camera negative pairs.
For person Re-ID, alleviating cross-camera intra-class variations is significant but ignored in MCNL.
To this end, we propose cross-camera feature prediction to mine cross-camera self supervision information by exploring the relations between transformed fake cross-camera positive pairs to help camera-invariant representation learning.
%By contrast, MCNL ignores the instance-level relations in the feature distribution.

\subsection{Self-Supervised Learning}

Self-supervised learning has gained popularity due to its ability to avoid the cost for annotation of large-scale datasets. It can adopt self-defined pseudo labels as supervision for training. Existing self-supervised learning methods can be categorized into three general types. First, generative self-supervised learning \cite{dinh2014nice,dinh2016density,kingma2018glow,razavi2019generating} aims to train an encoder to encode input $\mathbf{x}$ into a hidden representation $\mathbf{z}$ and a decoder to reconstruct $\mathbf{x}$ from $\mathbf{z}$.
Contrastive self-supervised learning \cite{caron2020unsupervised,chen2020simple,chen2020exploring,grill2020bootstrap,he2020momentum,tian2020makes} aims to train an encoder by pulling the embeddings of the same sample with different data augmentation closer while pushing embeddings of other samples away. Adversarial self-supervised learning \cite{mathieu2015masked,larsson2016learning,ledig2017photo,donahue2019large} aims to generate fake samples by training a generator and distinguish them from real samples by a discriminator. %Recently, contrastive self-supervised learning has become dominant in computer vision. The main idea is that pulling the embedding of the same sample with different data augmentation together while pushing embedding of other samples away.
Currently, contrastive self-supervised learning has become dominant in computer vision.
However, data augmentation strategy in current self-supervised learning methods is not specific for person Re-ID task and cannot reflect cross-camera intra-class variations.
In our method, cross-camera feature augmentation is camera-aware augmentation in feature level, which can help learning camera-invariant features.
%Moreover, data augmentation in these works is in image level and time consuming, while our augmentation is in feature level and requires fewer computation costs.
%For example, SimCLR\cite{chen2020simple} augments each image in a min-batch with two different augmentation methods and then pulls them close in the latent space while pushing other samples away. For more nagative contrast, the batch size is quite large (i.e., 4096). However, batch size is limited by the GPU memory size, a feasible way is to maintain a memory bank for negative pairs sampling\cite{he2020momentum}. However, maintaining a memory bank can be computationally expensive to update the representations in the memory bank. To overcome the issue, BYOL\cite{grill2020bootstrap} propose a momentum encoder to substitute of memory bank and only consider positive pairs from the same sample while ignoring the negative samples. SimSiam\cite{chen2020exploring} further use a siamese network instead of momentum one with stop-gradient operation and a MLP predictor to avoid model collapsing. However, all of these works do not consider real positive pairs which contains useful information. To this end, instead of instance-based contrastive approach, SwAV\cite{caron2020unsupervised} not only make different views of the same sample together but also makes sure that all features that are similar to each other and form clusters together. 

\section{APPROACH}

\subsection{Formulation}

A set of person images is collected and denoted by $\mathcal{X} = \left\{\mathcal{X}^{c} \right\}_{c=1}^{C}$, where $c\in \{1,...,C\}$ denotes the camera label.
%and $\mathcal{X}^{c}$ is person images captured from camera $c$.
In the setting of intra-camera supervised (ICS) Re-ID, the image-label pairs of camera $c$ are denoted by $\mathcal{X}^{c} = \left\{ (\mathbf{x}_{i}^{c},y_{i}^{c})\right\}_{i=1}^{N_c}$, where $\mathbf{x}_i^c$ is image and $y_{i}^{c}$ is the corresponding intra-camera identity label and $N_c$ is the number of samples. In the real world applications in large-scale surveillance systems, both matching between adjacent scenes and matching between distant scenes are required.
We categorize ICS setting into intra-camera supervised Re-ID across adjacent scenes (ICS-AS) and intra-camera supervised Re-ID across distant scenes (ICS-DS).
In this work, we study the under-explored ICS-DS setting,
in which cross-camera positive pair does not exist in training data captured in distant scenes, i.e., $y_{i}^{c_1} \neq y_{j}^{c_2}$ for all $i,j$ and $c_1 \neq c_2$. In comparison, the ICS-AS setting is studied in most existing ICS Re-ID methods, in which training data with a large number of underlying cross-camera positive pairs are captured in adjacent scenes, i.e., there exist $i,j$ and $c_1 \neq c_2$ that satisfy $y_{i}^{c_1} = y_{j}^{c_2}$. %$\exists i,j \in N_{c_1},N_{c_2}$ and $c_1 \neq c_2$ that satisfy $y_{i}^{c_1} = y_{j}^{c_2}$.

%denoted as $\left\{\mathbf{X}_{1_{c}}^{c}, \mathbf{X}_{2_{c}}^{c}, \ldots, \mathbf{X}_{l_{c}}^{c} \right\}$. Each entry in $\mathbf{X}_c$ is an image set belonging to the same category, i.e., $\mathbf{X}_{l_{c}}^{c} = \left\{x_{1}^{c}, x_{2}^{c}, \ldots, x_{n}^{c} \right\}$. 

%As assumed in ICS-AS, the tracking label in $\mathbf{X}_{l_{c}}^{c}$ is known, i.e, we know the images $\left\{x_{1}^{c}, x_{2}^{c}, \ldots, x_{n}^{c} \right\}$ share the same identity. In ICS-AS, cross-camera tracklet association is done to mine underlying positive pairs. While ICS-DS person Re-ID assumes a cross-camera unpaired training set, i.e., there is no positive pairs in $\mathbf{X}_{c_{1}}$ and $\mathbf{X}_{c_{2}}$.
We aim to train a feature extractor $H(\cdot ; \mathbf{\Theta})$ parameterized by $\mathbf{\Theta}$ on $\mathcal{X}$ to extract discriminative feature $\mathbf{f}_i^c = H( \mathbf{x}_{i}^{c} ; \mathbf{\Theta}) \in \mathbb{R}^{d}$ for image $\mathbf{x}_{i}^{c}$ to match. % guarantee the feature of each query $\boldsymbol{f}_{q}^{c_q} = H( x_{q}^{c_q} ; \mathbf{\Theta}) \in \mathbb{R}^{d}$ share more similarity with cross-camera pairs $\boldsymbol{f}_{g}^{c_g} = H( x_{g}^{c_g} ; \mathbf{\Theta}) \in \mathbb{R}^{d}$ (i.e., $c_g \neq c_q$) than with other images in the gallery with cross-camera unpaired training set.

\subsection{Cross-Camera Feature Prediction}

\subsubsection{\textbf{Cross-Camera Feature Augmentation}}

For person Re-ID task, the learning objective of feature extractor $H(\cdot ; \mathbf{\Theta})$ is %loss is which focuses on the relation of pair-wise samples is commonly employed for optimizing the  formulated as:
\begin{equation}
    \mathcal{D}(\mathbf{f}_{a}^{c_{a}}, \mathbf{f}_{p}^{c_{p}}) < \mathcal{D}(\mathbf{f}_{a}^{c_{a}}, \mathbf{f}_{n}^{c_{n}}),
    \label{11}
\end{equation}
where $\mathcal{D}$ is a distance metric, $\mathbf{f}_{a}^{c_{a}}, \mathbf{f}_{p}^{c_{p}}, \mathbf{f}_{n}^{c_{n}} $ are anchor feature from camera $c_a$, positive feature from camera $c_{p}$ and negative feature from camera $c_n$, respectively. The objective requires the distance between positive pairs to be smaller than that between negative pairs.
With the distance between cross-camera positive pair $\mathbf{f}_{a}^{c_{a}}, \mathbf{f}_{p}^{c_{p}}$ ($c_a \neq c_p$) minimized, cross-camera scene variations can be eliminated, which is helpful for learning discriminative representation.

However, in the ICS-DS setting, there is nearly no cross-camera positive pair in training data.
For an anchor feature $\mathbf{f}_{a}^{c_{a}}$, we can hardly find a $\mathbf{f}_{p}^{c_{p}}$ of a different camera ($c_a \neq c_p$), which makes it difficult to model intra-class cross-camera scene variations.
%With only intra-camera positive pairs $\boldsymbol{f}_{a}^{c_{a}}, \boldsymbol{f}_{p}^{c_{p}}$ ($c_a = c_p$), 
For ICS-DS Re-ID, Zhang \emph{et al.} \cite{zhang2020single} have experimentally shown that traditional metric learning loss results in dramatic performance degradation with a cross-camera unpaired training set.
How to establish the relation of cross-camera unpaired data is the key.
%In ICS-AS setting, there are cross-camera positive pairs, i.e., $c_a \neq c_p$. Pulling them together can learn camera-invariant representation. However, in ICS-DS setting, all samples of each class share the same camera information, i.e., $c_a = c_p$. In this case, only within-camera positive pairs are pulled close while cross-camera cues are not provided.

To overcome this problem, we replace the missing real cross-camera positive pair $\mathbf{f}_{a}^{c_a},\mathbf{f}_{p}^{c_p}$ ($c_a \neq c_p$) with fake cross-camera positive pair $\mathbf{f}_{a}^{c_a},\mathbf{\hat{\mathbf{f}}}_{p}^{c_p}$ ($c_a \neq c_p$). Fake feature $\mathbf{\hat{\mathbf{f}}}_{p}^{c_p}$ is obtained  by
\begin{equation}
\label{eq:augment}
\mathbf{\hat{\mathbf{f}}}_{p}^{c_p}  = \operatorname{Augment}(\mathbf{f}_{a}^{c_a}, c_p),
\end{equation}
where $\operatorname{Augment}$ is a transformation for feature $\mathbf{f}_{a}^{c_a}$ from camera $c_a$ to camera $c_p$. %to obtain a fake feature $\mathbf{\hat{\mathbf{f}}}_{p}^{c_p}$.
With the transformed fake feature, the learning objective in Eq. (\ref{11}) becomes:
\begin{equation}
    \mathcal{D}(\mathbf{f}_{a}^{c_{a}}, \hat{\mathbf{f}}_{p}^{c_p}) < \mathcal{D}(\mathbf{f}_{a}^{c_{a}}, \mathbf{f}_{n}^{c_{n}}).
\end{equation}

To realize this learning objective, we aim to design a transformation function and further design loss functions to explore the relation between transformed features, which can be regarded as scheme of mining self supervision information. 
Variations of lighting, viewpoint and background lead to camera-specific image style for Re-ID.
Since person images captured from different cameras are of different image styles and the image styles are camera-specific, we assume the feature distribution of each camera follows a camera-specific Gaussian distribution,
which is denoted as $\left\{ {N}(\boldsymbol{\mu}_{c},\boldsymbol{\sigma}_{c}^{2}) \right\}_{c=1}^{C}$.
%For features of all cameras, we assume they follow a global Gaussian distribution ${N}(\mu_{g},\sigma_{g}^{2})$.

%we do augmentation in the feature-level through a cross-camera augmentation module which is expected to generate cross-camera feature views which simulate features under different camera styles. %our CCP is expected to generate $C$ augmented cross-camera feature views which simulate features under different camera styles as shown in Fig. \ref{predictor} to capture more useful cross-camera cues and alleviate the cost for computation. 
We note that Batch Normalization (BN) layer \cite{ioffe2015batch} can estimate the moment statistics (i.e., mean and variance) of feature distribution and normalize features to normal distribution.
%to optimize under the global distribution $ \mathbf{N}(\mu_{g},\sigma_{g}^{2})$.
To realize the transformation, we propose a cross-camera feature augmentation module that consists of $C$ camera-specific Batch Normalization (CSBN) layers \cite{zhuang2020rethinking}, each of which estimates the camera-specific moment statistics $\left\{\boldsymbol{\mu}_{c},\boldsymbol{\sigma}_{c}^2 \right\}$.
%We mine self supervision information from camera-specific feature distribution.
Given a set of features of camera $c$ denoted as $\left\{ \mathbf{f}_{i}^{c}\right\}_{i=1}^{N_c}$, CSBN estimates camera-specific mean and variance by
\begin{equation}
    \boldsymbol{\mu}_{c} = \frac{1}{ N_c} \sum_{i=1}^{N_c} \mathbf{f}_{i}^{c},\quad {\boldsymbol{\sigma}_{c}^{2}} =  \frac{1}{ N_c} \sum_{i=1}^{N_c} (\mathbf{f}_{i}^{c} - \boldsymbol{\mu}_{c})^2,
\end{equation}
where $\boldsymbol{\mu}_{c}$ and $\boldsymbol{\sigma}_{c}$ are the camera-specific mean and variance.
%$\mu_{g}$ and $\sigma_{g}$ of global feature distribution can be estimated similarly.

Given a feature $\mathbf{f}_i^c$ of camera $c$, we transform it to a fake feature in the camera-specific distribution of a different camera $c'$ by
\begin{equation}
    \hat{\mathbf{f}}_{i}^{c'} = \operatorname{Augment}(\mathbf{f}_{i}^{c},c') = \gamma \frac{\mathbf{f}_{i}^{c}-\boldsymbol{\mu}_{c'}}{\sqrt{\boldsymbol{\sigma}_{c'}^{2}+\epsilon}} - \beta,
\end{equation}
where $\gamma$ and $\beta$ are two parameters learned during training. $\epsilon$ is a small value to keep stability.

\begin{figure}
\includegraphics[width=7.5cm]{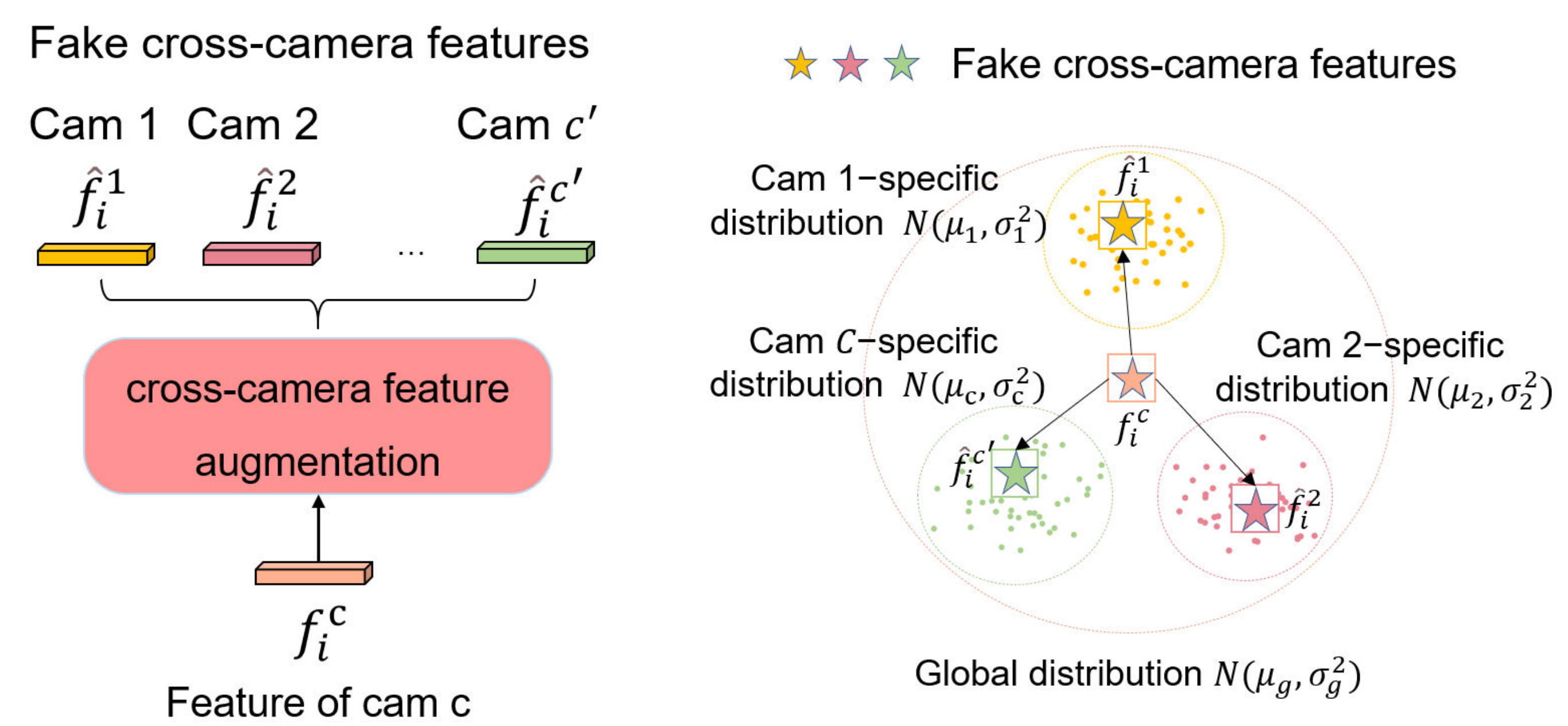}
\caption{Cross-Camera feature augmentation.
%(a). Augment feature to cross-camera features by a cross-camera augmentation module. (b). Cross-camera features-based self-supervised learning. $C$ augmented features of one feature are pulled together in the embedding space.
}
\label{predictor}
\vspace{-0.4cm}
\end{figure}

It is called {\em cross-camera feature augmentation} as shown on the left of Figure \ref{predictor}.
After this transformation, the feature $\mathbf{f}_i^{c}$ of camera $c$ is mapped to the distribution of camera $c'$ as shown on the right of Figure \ref{predictor}.
The relation between feature $\mathbf{f}_i^{c'}$ and feature $\mathbf{f}_i^c$ contains self supervision information of camera-specific feature distribution, which requires to be further mined.

\vspace{0.1cm}\noindent\textbf{Discussion.}
%Compared with CBN \cite{zhang2020rethinking,zhuang2021camera} for generalizable Re-ID, we apply camera-specific batch normalization (BN) in a different way for different purpose.
To model camera-specific feature distribution, both CBN \cite{zhuang2020rethinking,zhuang2021camera} for generalizable Re-ID and CSBN in our method exploit camera-specific batch normalization (BN) components, which are applied in different ways for different purposes.
In CBN \cite{zhuang2020rethinking,zhuang2021camera}, camera-specific BN operates on samples of the corresponding camera for distribution alignment; in our method, camera-specific BN operates on samples of all cameras to augment cross-camera fake features for further mining cross-camera self supervision information that is ignored by CBN \cite{zhuang2020rethinking,zhuang2021camera}.

\vspace{-0.1cm}

%and generate $\left\{ \hat{\boldsymbol{f}}_{a}^{c} \right\}_{c=1}^{C}$ with a cross-camera augmentation module as shown in Figure \ref{predictor}.
%We further generate fake cross-camera features $\left\{ \hat{\boldsymbol{f}}_{i}^{c} \right\}_{c=1}^{C}$ with inputting each feature $\boldsymbol{f}_{i}^{c_i}$ into all these $C$ camera-specific batch normalization layers and normalize it with each camera-specific statistics as shown in Figure \ref{predictor} :

%Specifically, as shown in Fig. \ref{predictor} (a) for each intermediate feature map $f_i \in \mathbb{R}^{C \times H \times W}$ where $C, H, W$ is the channel, height, weight of the feature map respectively. With CCP inserted into some special positions of the network, we further input the intermediate feature map $f_a \in \mathbb{R}^{C \times H \times W}$ into these $C$ CSBN layers and output $C$ cross-camera features $  f_{a}^{\hat{c}}, \hat{c} \in \left\{1,2, \ldots, C \right\} $ which encode the information of the same identify but under different camera distribution as shown in Fig. \ref{predictor} (b). With these cross-camera features, we then modify the metric learning loss as:where real cross-camera positive pairs $f_{p}^{c_{p}} $ are replaced with cross-camera features $f_{p}^{\hat{c}}$ to provide cross-camera cues for camera-invariant representation learning.

\begin{figure*}
    \centering
    \includegraphics[width=\textwidth]{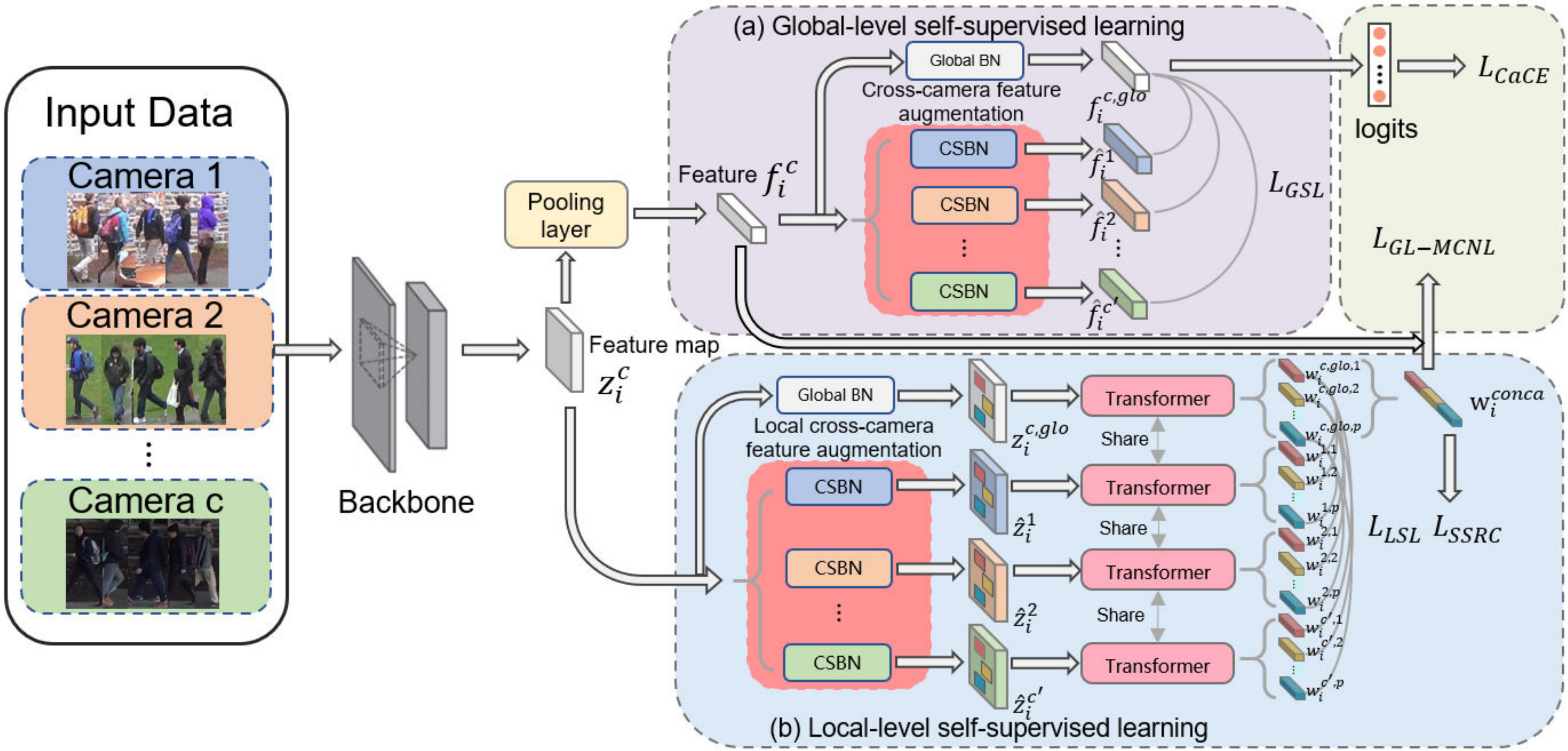}
    \caption{An overview of our framework.
    The cross-camera feature prediction scheme for mining cross-camera self supervision information is composed of (a) global-level self-supervised learning and (b) local-level self-supervised learning. In (a), GSL is designed to minimize the distances between features of the same intra-camera identity label, including ${\mathbf{f}}_i^{c,glo}$ output by global BN and the corresponding cross-camera features $ \hat{\mathbf{f}}_{i}^{c}, c \in (1,2,\ldots,C)$. In (b),  LSL is designed to minimize the distance between features $w_{i}^{c,glo,p}$ and ${w}_{i}^{c',p}$ of each local region.
    SSRC loss is a regularization term for guiding multiple local region learning.
    Finally, camera-aware cross entropy loss and GL-MCNL loss are employed for representation learning with intra-camera supervision.}
    \label{overrall1}
\end{figure*}

\subsubsection{\textbf{Predicting Cross-Camera Fake Feature}}

The relation between transformed cross-camera features contain self supervision information of camera-specific distribution.
To mine cross-camera self supervision information for camera-invariant representation learning, we propose a {\em cross-camera feature prediction} scheme as shown in Figure \ref{overrall1}.

%In this section, our goal is to detail how to use augmented cross-camera features for self-supervised learning. For most of existing Re-ID methods, whole image-based features are commonly employed for optimization. However, Yang \emph{et al}\cite{yang2019patch} proposes that pulling the features of the similar person images together would blur identity information of person image, making it ineffective to distinguish the similar images of different identities, so optimizing in the local-level is also very important and is complementary with global-level optimization. To this end, we propose a self-supervised learning method both in global-level and local-level with augmented cross-camera features.
%\subsubsection{\textbf{Global-level self-supervised learning}}
%In this section, we aim to use the cross-camera features to guide the global-level self-supervised learning.

We assume the features of all cameras follow a global Gaussian distribution.
Similar to the camera-specific BN, we apply a BN layer called global BN on $\mathbf{f}_i^c$ to obtain a normalized feature $\mathbf{f}_i^{c,glo}$, which is also the feature for inference in the testing stage.
For each sample $\mathbf{x}_i$ as input, we can obtain $C+1$ views of features, including a normalized feature $\mathbf{f}_{i}^{c,glo}$ and fake cross-camera features $ \left\{\hat{\mathbf{f}}_{i}^{c'} \right\}_{c'=1}^{C}$. Since all of them are encoded from the same image $\mathbf{x}_i$, we regard these features of different camera-specific views share the same identity.
For a fake cross-camera positive pair of feature $\mathbf{f}_{i}^{c}$ and feature $\hat{\mathbf{f}}_{i}^{c'}$, we expect that $\mathbf{f}_{i}^{c}$ can be used to predict $\hat{\mathbf{f}}_{i}^{c'}$ by mapping $\mathbf{f}_{i}^{c}$ to $\mathbf{f}_{i}^{c,glo}$ in a global distribution, which can be formulated by
\begin{equation}
   \mathcal{L}= \frac{1}{N}\frac{1}{C} \sum_{i=1}^{N}\sum_{c'=1}^{C} \mathcal{D}\left(\mathbf{f}_{i}^{c,glo}, \text{stop-gradient}(\hat{\mathbf{f}}_{i}^{c'})\right),
    \label{original}
\end{equation}
where $\hat{\mathbf{f}}_{i}^{c'}$ is regarded as the prediction target so that stop-gradient is applied to it.
%$\boldsymbol{f}_{i}$ is the feature after the global BN which estimates the global statistics of whole training set.
%$\hat{\boldsymbol{f}}_{i}^{c}$ is the cross-camera feature output by $c$-th head of cross-camera augmentation module which estimates the statistics of local training set from camera $c$,
$\mathcal{D}$ is cosine distance.

However, the formulation in Eq. (\ref{original}) only takes fake features transformed from a single feature into account, while intra-camera supervision is ignored, which is crucial for alleviating intra-class variations.
Thus, we also minimize the distances between $\mathbf{f}_i^{c,glo}$ and cross-camera features of the same intra-camera identity with $\mathbf{f}_i^{c,glo}$, in order to alleviate the effect of both cross-camera scene variations and intra-class variations.
%within a mini-batch into designing of loss. Specifically,
%let the samples in a mini-batch be denoted as $\left\{(x_{i},y_i)  \right\}_{i=1}^{N}$, where $N$ is the batch size. %The positive pair of $(x_i, y_i)$ in the mini-batch is denoted as $(x_j, y_j)$, where $ y_j = y_i$.
%All samples of the same identity as well as their augmented cross-camera features should be compact in the embedding space to capture cross-camera cues and intra-class variations.
%For each feature $f_{i}$ input into the cross-camera augmentation module, we get $C$ cross-camera features. 
%So for $P+1$ samples from the same class, we have $c \times (P_{i}+1)$ CCFs for the class. We split these features into camera-special ones according to the head output of CCP, i.e., we stack the features outputted by BNNeck as $\left\{ f_{1}^{a},f_{2}^{a},\ldots,f_{P_{i}+1}^{a} \right\}$, the features outputted by $\text{BN}_c$ as $\left\{ F_{1},F_{2},\ldots,F_{P_{i}+1} \right\}$ where $F_{P_{i}+1}$ is a feature set include $c$ CCFs for sample $f_{P_{i}+1}$ denoted as $\left\{f_{P_{i}+1}^{1},f_{P_{i}+1}^{2},\ldots,f_{P_{i}+1}^{c}\right\}$. We enforce these features of the same class to be compact in the embedding space. 
To achieve this, %with the feature extracted from the person image,
we formulate a \textbf{G}lobal \textbf{S}elf-supervised \textbf{L}oss (GSL) as follow:

\begin{equation}
    \mathcal{L_{\text{GSL}}}=\frac{1}{N}\frac{1}{C}\frac{1}{| \mathcal{P}_{i} |}\sum_{i=1}^{N}\sum_{c'=1}^{C}\sum_{j \in \mathcal{P}_{i}} \mathcal{D}\left(\mathbf{f}_{i}^{c,glo}, \text{stop-gradient}(\hat{\mathbf{f}}_{j}^{c'})\right),
\end{equation}
where $\mathcal{P}_{i}$ denotes the set of indices of samples that share the same identity with $\mathbf{x}_i$ (i.e., $y_i = y_j$) and $| \mathcal{P} | $ is the cardinality of $\mathcal{P}$.

This is called {\em cross-camera feature prediction}. By minimizing the distances between the fake cross-camera positive pairs, self supervision information in camera-specific feature distribution is mined and used to guide camera-invariant feature learning.

%They should be pulled together in the embedding space to align the feature distribution between camera-specific and whole training set.
%Regular distribution align loss optimize in the global-level, i.e., they enforce the global statistics (e.g., mean and variance) among domains to be similar which ignore the relations between sample pairs. Our method aims to mine self-supervised information to simulate pair-wise instances aligning which achieve better performance.we regard mapping a feature to its corresponding generated cross-camera fake feature as cross-camera feature prediction
%we regard mapping a feature to its corresponding augmented cross-camera feature as cross-camera feature prediction and formulate the loss with a stop-gradient operation as: %as in Simsiam\cite{chen2020exploring} as:

\vspace{0.1cm}\noindent \textbf{Discussion.}
The commonly used distribution alignment methods, such as MMD \cite{mmd} and CORAL \cite{coral} and HHL \cite{zhong2018generalizing}, model the relation between different domains by minimizing distribution distance, while instance-level relation is ignored.
By contrast, our method models the relation between different domains by fake cross-camera positive pairs in instance level, which is exploiting self supervision information mined from domain-specific distributions.
Such instance-level modeling is significant for alleviating intra-class variations and thus is more suitable for Re-ID task.
%in  in the global-level, i.e., they minimize the distance among distribution while ignore the instance-level relation in different distribution, our method aims to mine self-supervised information to align the distribution in instance-level.

\subsection{Global-Local Cross-Camera Feature Prediction Scheme}
The global self-supervised loss exploits global-level features extracted from the whole image for self-supervised learning.
To extract more fine-grained features to complement global-level self-supervised learning, we further propose a global-local cross-camera feature prediction scheme, which aggregates fine-grained local-level feature with global-level feature.
We expect that our model can automatically localize local regions from the feature map based on contextual information and then perform self-supervised learning in local level to mine fine-grained self supervision information.
To this end, we employ a Transformer \cite{vaswani2017attention} which can effectively model contextual information for local self-supervised learning in our cross-camera feature prediction scheme. 

\vspace{0.1cm}\noindent \textbf{Local Feature Extraction by Transformer.}
Motivated by DETR \cite{carion2020end}, we apply a transformer for extracting local-level features of multiple local regions on feature map $\mathbf{z}_{i} \in \mathbb{R}^{d \times H \times W}$ extracted by the backbone model.
The transformer consists of an encoder $E(\cdot;\mathbf{\Theta}_E)$ for learning contextual information and a decoder $D(\cdot;\mathbf{\Theta}_D)$ for localizing region of interest based on contextual information for local-level feature extraction.
%For encoder, a $1 \times 1$ convolution is employed to reduce the channel dimension of the feature map output by the backbone to dimension $d$, which is denoted as $\boldsymbol{z}_{i} \in \mathbb{R}^{d \times H \times W}$.
We resize the feature map $\mathbf{z}_{i} \in \mathbb{R}^{d \times H \times W}$ to obtain $\mathbf{z}_i \in \mathbb{R}^{d \times HW}$ as the input of encoder.
%Each encoder layer consists of a multi-head self-attention module and a feed forward network. Since we hope the encoder can focus on local regions with rich semantic information, we supplement it with learnable positional encodings which are added to the input of each attention layer. 
The output embedding of the encoder $\mathbf{z}_{i}^{e}\in \mathbb{R}^{ d \times HW}$ is
\begin{equation}
    \mathbf{z}_{i}^{e} = E(\mathbf{z}_i;\mathbf{\Theta}_E ).
\end{equation}
The decoder takes $\mathbf{z}_{i}^{e}$ as input and finally outputs $P$ local-level features:
\begin{equation}
    \left\{ \mathbf{w}_{i}^{p}\right\}_{p=1}^{P} = D(\mathbf{z}_{i}^{e};\mathbf{\Theta}_D).
\end{equation}
%also follows the standard architecture of transformer which transforms $N$ embeddings of size $d$ using multi-head self-attention machanisms. 
%Since the decoder is expected to produce different results with the permutation-invariant property, the $N$ input embeddings must be different, object queries which are learnable to encode the positional information are added to the input of each attention layer. Finally, the decoder outputs $P$ features $ \boldsymbol{w}_{i}^{p} = D(\boldsymbol{z_2};\Theta_D) \in \mathbb{R}^{256} , p \in (1,2,\ldots,P) $ which act as local region features in our work.
As for the detailed architecture for the transformer, we follow the design in DETR \cite{carion2020end}.

\vspace{0.1cm}\noindent \textbf{Local-level Self-Supervised Learning.}
Since local-level feature contains more fine-grained information, to exploit local-level feature for self-supervised learning, we propose a \textbf{L}ocal-level \textbf{S}elf-supervised \textbf{L}oss (LSL). %as a complement of the global-level self-supervised loss.
Similar to the model architecture used in global-level self-supervised learning,
we apply another local cross-camera feature augmentation module and global BN after the backbone in the feature map level as shown in Figure \ref{overrall1} (b).
The feature map $z_{i}^{c}$ output by the backbone is further input to the global BN and the local cross-camera augmentation module to get a normalized feature map $ \mathbf{z}_{i}^{c,glo}$ and cross-camera feature maps $\left\{ \mathbf{\hat{z}}_{i}^{c'}\right\}_{c'=1}^{C}, i \in (1,2,\ldots,N)$, respectively.
We assimilate the idea of global-level self-supervised learning into local-level learning. Specifically,  We input $\mathbf{z}_{i}^{c,glo} $ and $\left\{ \mathbf{\hat{z}}_{i}^{c'}\right\}_{c'=1}^{C}$, $ i \in (1,2,\ldots,N)$  into the Transformer to extract local-level features. For each feature map $\mathbf{z}_{i}^{c}$ as input, $P$ local-level features $\left\{\mathbf{w}_{i}^{c,p} \right\}_{p=1}^{P}$ are output by the transformer.
Similar to the GSL, we enforce each local-level feature  $\mathbf{w}_{i}^{c,glo,p}$ of $\mathbf{z}_{i}^{c,glo}$ and $\mathbf{w}_{i}^{c',p}$ of corresponding cross-camera feature maps $\mathbf{\hat{z}}_{i}^{c'}$ for each sample to be pulled close in the embedding space:
\begin{equation}
    \mathcal{L}_{\text{LS}}=\frac{1}{N}\frac{1}{C}\frac{1}{P}\sum_{i=1}^{N}\sum_{c'=1}^{C}\sum_{p=1}^{P}\mathcal{D}\left(\mathbf{w}_{i}^{c,glo,p}, \text{stop-gradient}(\mathbf{w}_{i}^{c',p})\right).
\end{equation}

To guide multiple local region learning, we introduce a shared-specific region constraint (SSRC) loss as a regularization term, which consists of a specific loss and a shared loss as follows.

On the one hand, we expect that different regions can focus on different parts with large appearance variations to extract part-specific features. To this end, 
We employ specific loss to push local-level features $\left\{\mathbf{w}_{i}^{c,glo,p} \right\}_{p=1}^{P}, i \in (1,2,\ldots,N)$ away from each other formulated as follow: %which can effectively avoid all local regions from focusing on the single region.
\begin{equation}
    \mathcal{L}_{specific}=-\sum_{i=1}^{N}\sum_{p=1}^{P}\log \frac{ \operatorname{exp} \left( s\left(\mathbf{w}_{i}^{c,glo,p}, \mathbf{w}_{i}^{c,glo,p}\right) \right) }{  \sum_{q = 1}^{P}\operatorname{exp} \left( s\left(\mathbf{w}_{i}^{c,glo,p}, \mathbf{w}_{i}^{c,glo,q}\right) \right)},
\end{equation}
where $s(\cdot)$ measures the cosine similarity between two features.

On the other hand, we also expect each region can focus on some shared patterns of different persons to avoid attending to person-irrelevant regions. 
%them to be shared on cross instance-level to capture similar representations on different instances to facilitate fine-grained local-level self-supervised learning.
We regard region index $p$ as class label and employ Softmax cross entropy loss to classify the feature as follow: %push local-level features close to corresponding representation center:
\begin{equation}
    \mathcal{L}_{share}=-\sum_{i=1}^{N}\sum_{p=1}^{P} \mathbf{C}_{i}^{p}\operatorname{log}\mathbf{q}_{i}^{c,glo,p},
\end{equation}
where $\mathbf{C}_{i}^{p}$ is a one-hot vector representing the index $p$  and $\mathbf{q}_{i}^{c,glo,p} $ is the output logit of fully connected layer with softmax operation.
In summary, The SSRC loss is formulated as:
\begin{equation}
    \mathcal{L}_{SSRC}=\mathcal{L}_{\mathrm{specific}} + \mathcal{L}_{\mathrm{share}}.
\end{equation}

Finally, the total loss for local-level self-supervised learning is:
\begin{equation}
    \mathcal{L}_{\mathrm{LSL}} = \mathcal{L}_{\mathrm{LS}} + \mathcal{L}_{\mathrm{SSRC}}.
\end{equation}

\subsection{An ICS-DS Re-ID Framework}

To learn from intra-camera identity labels in the ICS-DS setting, we adopt the commonly used classification loss and metric learning loss and further aggregate cross-camera self supervision and intra-camera supervision into our ICS-DS Re-ID framework. %Our method is compatible with regular loss for person Re-ID task. Given the intra-camera tracking annotation, we can adopt the commonly used classification loss and metric learning loss to train the model. Specifically, our framework is designed upon MCNL\cite{zhang2020single} with \textbf{C}amera-\textbf{a}ware \textbf{C}ross-\textbf{E}ntropy loss (CaCE).

\vspace{0.1cm}\noindent\textbf{Global-local MCNL.}
For discriminative feature learning, we apply a metric learning loss MCNL \cite{zhang2020single} that is designed for ICS-DS setting.
MCNL loss exploits intra-camera pairs and cross-camera negative pairs for learning from intra-camera supervision information, which is complementary to our cross-camera feature prediction that exploits fake cross-camera positive pairs to learn from cross-camera self supervision information.

%Since we have both global-level feature and local-level feature in our method, 

For each feature map $\mathbf{z}_{i}^{c,glo}$, we concatenate the local-level features $\left\{\mathbf{w}_{i}^{c,glo,p} \right\}_{p=1}^{P}$ output by Transformer to obtain a combined local-level feature $\mathbf{w}_{i}^{conca}$. %MCNL\cite{zhang2020single} on both concatenated local-level feature $\left\{  \mathbf{w}_{i}^{conca}\right\}_{i=1}^{N}$ and global-level feature $\left\{  \boldsymbol{f}_{i}^{c}\right\}_{i=1}^{N}$.
We apply MCNL loss on both global-level feature $\mathbf{f}_{i}^{c}$ and local-level feature $\mathbf{w}_{i}^{conca}$ to formulate a global-local MCNL (GL-MCNL) loss formulated by:
\begin{equation}
\begin{aligned}
\mathcal{L}_{GL-MCNL}=& \frac{1}{N} \sum_{* \in \{glo,loc\}} \sum_{i=1}^{N}\left[m_{1}+\operatorname{dist}_{+,\text{intra}}^{*,i}-\operatorname{dist}_{-, \text {cross }}^{*,i}\right]_{+} \\
&+\left[m_{2}+\operatorname{dist}_{-, \text {cross}}^{*,i}-\operatorname {dist}_{-, \text {intra }}^{*,i}\right]_{+}, 
\end{aligned}
\end{equation}
where $[\mathbf{z}]_{+}=\operatorname{max}(\mathbf{z},0)$, $\operatorname{dist}_{+}^{*,i}$ denotes distance between anchor and within-camera hard positive pair. $\operatorname{dist}_{-, \text {cross }}$ denotes distance between anchor and cross-camera hard negative pair. $\operatorname {dist}_{-, \text {intra}}^{*,i}$ denotes distance between anchor and intra-camera hard negative pair. $* \in \left\{glo,loc\right\}$ denotes distance calculated by global-level feature and local-level feature, respectively.

%For local-level MCNL (L-MCNL) and global-level MCNL (G-MCNL), the $\operatorname{dist}_{+}^{i}$ is Euclidean distance between concatenated features $\boldsymbol{t}_i$ and global features $\boldsymbol{f}_i$ of anchor and that of within-camera hard positive respectively. $\operatorname{dist}_{-, \text {other }}^{i}$ is Euclidean distance between concatenated features $\boldsymbol{t}_i $ and global features $\boldsymbol{f}_i$ of anchor and that of cross-camera hard negative. $\operatorname{dist}_{-, \mathrm{same}}^{i}$ is Euclidean distance between concatenated features $\boldsymbol{t}_i$ and global features $\boldsymbol{f}_i$ of anchor and that of within-camera hard negative.

\vspace{0.1cm}\noindent\textbf{Camera-aware Cross Entropy Loss.}
We assimilate the main idea of MCNL to formulate a new camera-aware cross entropy (CaCE) loss. CaCE loss is formulated as:
\vspace{-0.1cm}
\begin{equation}
    \mathcal{L}_{\text{CE}}=-\sum_{i=1}^{C} \mathbf{p}_{i} \log \mathbf{q}_{i},
\end{equation}

where $\mathbf{q}_{i}$ is the output logits of sample $x_i$ representing the possibilities of classes and $\mathbf{p}_{i}$ is the ground truth soft label formulated as:
%In cross entropy loss with label smooth, except for the class of ground truth, the possibility of other classes are all set as $\varepsilon /(C-1)$ which means the sample is constrained to keep the similar distance with class centers of other classes which is contradictory with the thought of MCNL. To make Cross Entropy Loss more compatible with MCNL, we slightly change the label initial method formulated as:
\begin{equation}
    \mathbf{p}_{i}=\left\{\begin{array}{ll}
    1-\varepsilon & \text { if } i=j \\
    -\varepsilon /(K_{e}-1) & \text { if } i \neq j \land c_i \neq c_j\\
    \varepsilon /(K_{n}-1) & \text { if }  i \neq j \land c_i = c_j
    \end{array}\right.,
\end{equation}
where $c_i$ is the camera label of sample $\mathbf{x}_i$. $K_{e}$ is the number of classes whose camera label is the same with $c_i$  and $K_{n}$ is the number of class whose camera label is different from $c_i$. $\varepsilon$ is set as 0.1. 

Finally, our framework is learned with joint guidance of cross-camera self supervision and intra-camera supervision by
\begin{equation}
    \mathcal{L}_{\mathrm{CCFP}} = \mathcal{L}_{\mathrm{CaCE}} + \mathcal{L}_{\mathrm{GL-MCNL}} + \mathcal{L}_{\mathrm{GSL}} + \mathcal{L}_{\mathrm{LSL}}.
\end{equation}

During testing, we only use features $\mathbf{f}_{i}^{c}$ output by global BN in Figure \ref{overrall1} (a) for testing, which can save computation costs.  

\section{Experiments}

\subsection{Datasets and Evaluation Protocol}

To evaluate the effectiveness of our model, we conducted experiments in the ICS-DS setting on three benchmark person Re-ID datasets , Market-SCT, DukeMTMC-SCT and MSMT17-SCT, which are modified from Market-1501 \cite{zheng2015scalable}, DukeMTMC \cite{duke} and MSMT17 \cite{wei2018person} following the data split in single camera training (SCT) \cite{zhang2020single}, because existing datasets do not meet the requirement of having cross-camera unpaired training data. 
Compared with the original dataset, we randomly select images of one camera for each identity to construct the training set and keep the original testing data and testing protocols unchanged.
The details of datasets used in our experiments are shown in Table \ref{detail}. The evaluation metrics are the Rank-k accuracy
and the mean average precision (mAP)\cite{2015_ICCV_MARKET}.
%However, intra-camera supervised person Re-ID across distant scenes assume a cross-camera unpaired data. Since the original training set do not meet the requirement,  we first split these training sets into Market-SCT, Duke-SCT and MSMT-SCT respectively as did in SCT\cite{zhang2020single}. Specifically, we randomly choose one camera for each person and take those images of the person under the selected camera as training images while discard others. 
%For fair comparison, we took the Market-SCT and Duke-SCT split in MCNL\cite{zhang2020single}. Since they did not conduct experiments on MSMT17, we reconstructed MSMT into MSMT-SCT according to the aforementioned method of splitting. 

%Besides, due to a too absolute hypothesis that each person will not be captured in different cameras for SCT, we relax the hypothesis and reconstruct training set with certain percent overlap rate, i.e., we randomly choose some identities whose images will be cross-camera paired data but with a different identity labelled. We set the overlap rate as $5\% $ to $50\%$ with increasing by $5\%$ each time and carried out experiments on these datasets and prove our model is robust under the outliers.

\subsection{Implementation Details}

ResNet-50 \cite{he2016deep} pre-trained on ImageNet \cite{deng2009imagenet} was adopted as our backbone. The details of its structure followed the design in AGW \cite{ye2021deep}.
%A global BN layer and local cross-camera feature augmentation module were inserted after the backbone in the feature map level for local-level self-supervised learning.
For local-level feature extraction, a 1x1 convolution was used to reduce the channel dimension of the feature map $\mathbf{z}_{i}^{c}$ output by the backbone to a smaller dimension of 256 before it was input to Transformer.
The parameters of Transformer in local-level self-supervised learning followed DETR \cite{carion2020end}, except that the number of local regions $P$ was set as $12$ empirically.
The feature output by pooling layer was used for metric learning.
%After pooling, another global BN and cross-camera augmentation feature module were inserted for the global level self-supervised learning.
The data augmentation strategy followed the implementation in fast-reid \cite{he2020fastreid}.
The mini-batch size was 128.
As for mini-batch sampling strategy,
we randomly selected images from 2 cameras.
%and sample 16,8,8 identities for Market-SCT, Duke-SCT and MSMT-SCT for each selected cameras, respectively. 
We used Adam \cite{kingma2014adam} as our optimizer, and the learning rate was set to 0.0006 initially and decayed by 0.1 every 100 epochs in 300 epochs totally.
%All weights of the loss used in our framework were set as 1. 
%We removed its final fully connected layer and changed the stride of last residual block to 1. Non-local attention block was inserted into the backbone with the default setting from \cite{wang2018non} and using generalized-mean pooling (GeM)\cite{radenovic2018fine} which can extract the domain-specific discriminative features instead of Global Average Pooling (GAP) after the backbone. A global BN layer and a cross-camera augmentation module are inserted after the backbone in the feature map level for local-level self-supervised learning. Structure of Tranformer follows DETR\cite{carion2020end} with changing the number of local regions to $12$. The feature after Gem pooling layer is used for metric learning. After pooling, another global BN and cross-camera augmentation module are inserted for the global level self-supervised learning. The augmentation for training set follow the default setting in fast-reid\cite{he2020fastreid}. Each mini-batch contains 128 images with our proposed Bi-cameras sample strategy, we randomly select 2 cameras and sample 16,8,8 identities for Market-SCT, Duke-SCT and MSMT-SCT for each selected cameras respectively. We use the Adam\cite{kingma2014adam} as our optimization algorithm, and the learning rate was set to 0.0006 initially and decayed by 0.1 every 100 epochs with 300 epochs totally. All weights of the loss used in our framework are set to 1. 

\newcommand{\tabincell}[2]{\begin{tabular}{@{}#1@{}}#2\end{tabular}}
\begin{table}[]
\small
    \centering
    \caption{Details of datasets used in our experiments.}
    \begin{tabular}{c|c|c|c|c|c}
    \hline
         Dataset & Train &Train &Test & Test & cross-camera \\ &IDs & Images &IDs & Images  & paired data   \\ \hline
         Market-1501 & 751 &12936 &750  &15913  &True \\ \hline
         Market-SCT & 751 &3561 &750  &15913  &False \\\hline
         DukeMTMC & 702 &16522 &1110  &17661  &True \\\hline
         Duke-SCT & 702 &5993 &1110  &17661  &False \\\hline
         MSMT17 & 1041 &32621&3060&93820  &True \\\hline
         MSMT-SCT & 1041 &6645 &3060  &93820  &False \\ \hline
    \end{tabular}
    \label{detail}
\end{table}

\begin{table*}[]
\small
    \centering
    \begin{tabular}{c|c|c||cccc||cccc||cccc}
         \hline
          \multirow{2}*{Setting}& \multirow{2}*{Methods}& \multirow{2}*{Reference}& \multicolumn{4}{c||}{Duke-SCT}&   \multicolumn{4}{c||}{Market-SCT}&  \multicolumn{4}{c}{MSMT17-SCT} \\ \cline{4-15}
           &&  &R-1  &R-5  &R-10  &mAP  &R-1  &R-5  &R-10   &mAP  &R-1  &R-5  &R-10   &mAP  \\  \hline\hline
          \multirow{5}*{\textbf{ICS-AS}} &\textbf{ICS} &  & & & &&  & & & &- &- &- &- \\
          &$\text{TAULD}^{\S} $ \cite{tauld}  &ECCV 18  &61.7 &- &- &43.5  &63.7  &- &- &41.2  &- &- &- &- \\
          &$\text{TSSL}^{\S} $ \cite{tssl} &AAAI 20  &62.2 &- &- &38.5 &71.2  &- &- &43.3  &- &- &- &- \\ 
          &$\text{MTML}^{\S} $ \cite{zhu2019intra} &ICCV 19  &71.7  &- &86.9 &50.7 &85.3 &- &96.2 &65.2  &44.1 &- &63.9 &18.6 \\ 
          &$\text{UGA}^{\S}$ \cite{graph_asso}  &ICCV 19  &75.0  &- &- &53.3 &87.2 &- &- &70.3  &49.5 &- &- &21.7 \\ 
          &$\text{PCSL}^{\S} $ \cite{qi2020progressive} &TCSVT20  &71.7  &84.7 &88.2 &53.5  &87.0 &94.8 &96.6 &69.4  &48.3 &62.8 &68.6 &20.7 \\  \hline\hline

          \multirow{19}*{\textbf{ICS-DS}}
          &\textbf{Fully Supervised} & & & & &  & & & &  & & &  & \\
          &PCB \cite{sun2018beyond}  &ECCV 18  &32.7 &- &- &22.2  &43.5 &- &- &23.5  &- &- &-  &- \\
          &Suh's method \cite{suh2018part}  &ECCV 18 &38.5 &- &- &25.4  &48.0 &- &- &27.3  &- &- &-  &- \\
          &MGN-ibn \cite{MGN} &ACMMM' 18  &46.7 &59.8 &67.1 &32.6  &45.6 &61.2 &69.3 &26.6  &27.8 &38.6 &44.1  &11.7 \\
          &bagtricks \cite{luo2019bag} &CVPR 19 &54.2 &68.9 &76.7 &42.0 &54.0 &71.3 &78.4 &34.0  &20.4 &31.0 &37.2  &9.8\\
          &AGW \cite{ye2021deep} &TPAMI 21 &56.5 &71.0 &77.7 &43.9 &56.0 &72.3 &79.1 &36.6  &23.0 &33.9 &40.0  &11.1\\ \cline{2-15}
          &\textbf{Metric Learning} & & & & &  & & & &  & & &  & \\
          &Center Loss \cite{center}  &ECCV 16  &38.7 &- &- &23.3  &40.3 &- &- &18.5  &- &- &-  &- \\
          &A-Softmax \cite{Asoft}  &CVPR 17  &34.8 &- &- &22.9  &41.9 &- &- &23.2  &- &- &-  &- \\
          &ArcFace \cite{deng2019arcface}  &CVPR 19 &35.8 &- &- &22.8  &39.4 &- &- &19.8  &- &- &-  &-\\\cline{2-15}
          &\textbf{Self-supervised} & & & & &  & & & &  & & &  & \\
          &$\text{HHL}^{\dagger} $  \cite{zhong2018generalizing} &ECCV 18 & 50.3 & 64.0 & 70.3 & 33.0 &65.6 & 80.6 &86.8 & 44.8 & 31.4 &42.5 & 48.1 & 11.0 \\
          &SimSiam \cite{chen2020exploring} &CVPR 21  &28.1 & 43.2 & 51.3 & 19.7  &36.2 & 51.9  & 59.1 & 18.0  &2.8  & 5.9 & 8.4 & 1.2 \\\cline{2-15}
          &\textbf{Distribution alignment} & & & & &  & & & &  & & &  & \\
          &MMD \cite{mmd}  &JMLR 15  &74.1 &85.9 &90.2 &56.3  &67.7  &83.1 &88.2 &44.0  &42.2 &55.8 &61.4 &18.2 \\
          &CORAL \cite{coral} &ECCV 16  &76.8 &87.1 &90.0 &59.7  &76.2  &88.5 &93.0 &51.5  &42.6 &55.8 &61.5 &19.5 \\\cline{2-15}
          &\textbf{ICS} &  & & & &&  & & & &- &- &- &- \\
          &Precise-ICS \cite{wang2021towards} &WACV 21  &41.2 &57.9 &64.2 &25.9 &50.0  &67.5 &74.8 &31.2  &17.2 &28.4 &34.3 &6.7 \\
          &$\text{MCNL}^{\ddag} $ \cite{zhang2020single}   &AAAI 20  &67.1 &80.9 &84.7 &45.2  &67.0 &82.8 &87.9 &41.6  &26.6 &40.0 &46.4  &10.0 \\ \cline{2-15}
          
          &CCFP(ours)   &ours  &\textbf{80.3} &\textbf{89.0} &\textbf{91.9} &\textbf{64.5}  &\textbf{82.4} &\textbf{92.6} &\textbf{95.4} &\textbf{63.9} &\textbf{50.1} &\textbf{63.3} &\textbf{68.8}  &\textbf{22.2}  \\ \hline

    \end{tabular}
    \caption{Comparison with state-of-the-art methods on Duke-SCT, Market-SCT and MSMT17-SCT. R-k denotes Rank-k accuracy (\%) and mAP denotes mean average precision (\%). $\text{MCNL}^{\ddag} $ \cite{zhang2020single}: We used the code released by the authors and achieved higher performance than that reported in the paper.
    $\S$ TAULD \cite{tauld}, TSSL \cite{tssl}, MTML \cite{zhu2019intra}, $\text{UGA}$  \cite{graph_asso} PCSL \cite{qi2020progressive}: ICS Re-ID models trained with cross-camera paired data in the ICS-AS setting. $\text{HHL}^{\dagger} $: we modified unsupervised HHL as a intra-camera supervised version. Results in bold font denote the best performance in the ICS-DS setting.
    }

    \label{sota}
    %\vspace{-0.6cm}   
\end{table*}

\subsection{Comparison with State-of-the-Art Methods}

We evaluated several methods including the state-of-the-art intra-camera supervised methods, fully supervised learning methods, self-supervised learning methods, metric learning methods and distribution alignment methods. The results are shown in Table \ref{sota}.
%We outperformed the state-of-the-art ICS-DS Re-ID method MCNL\cite{zhang2020single} by a large margin.
%(including PCB\cite{suh2018part}, Suh's method\cite{sun2018beyond}, MGN\cite{MGN}, BoT\cite{luo2019bag} and AGW\cite{ye2021deep}), metric learning methods (including Center loss\cite{center}, A-Softmax\cite{Asoft} and Arcface\cite{deng2019arcface}), self-supervised learning methods (including Simsiam\cite{chen2020exploring}, and HHL\cite{zhong2018generalizing}), distribution alignment methods (including MMD\cite{mmd} and CORAL\cite{coral}) and intra-camera supervised methods (including TAULD\cite{tauld}, TSSL\cite{tssl}, MTML\cite{zhu2019intra}, UGA\cite{graph_asso}, PCSL\cite{qi2020progressive} and Precise-ICS\cite{wang2021towards}) as shown in Table \ref{sota}. We outperformed the state-of-the-art ICS-DS Re-ID method MCNL\cite{zhang2020single} by a large margin.

\vspace{0.1cm}\noindent \textbf{Intra-camera supervised (ICS) methods.} We categorize ICS Re-ID methods into ICS-AS methods and ICS-DS methods.

\vspace{0.1cm}\noindent \textbf{- ICS-AS methods.}
We compared with methods for intra-camera supervised Re-ID across adjacent scenes (ICS-AS) including TAULD \cite{tauld}, TSSL \cite{tssl}, MTML \cite{zhu2019intra}, UGA \cite{graph_asso}, PCSL \cite{qi2020progressive} and Precise-ICS \cite{wang2021towards}.
They were designed for small-scale surveillance systems and their training set included underlying cross-camera positive pairs. 
Since Precise-ICS \cite{wang2021towards} achieved best performance in the ICS-AS setting, we implemented it in the ICS-DS setting. For other methods, we used their results in the ICS-AS setting reported in their original papers.
Compared with ICS-AS setting, ICS-DS setting is more challenging without cross-camera paired training data, but our method still surpassed the ICS-AS Re-ID methods on Duke-SCT and MSMT-SCT.
Moreover, our method clearly outperformed Precise-ICS in the same ICS-DS setting.

\vspace{0.1cm}\noindent \textbf{- ICS-DS methods.}
We compared with the state-of-the-art ICS-DS Re-ID method MCNL \cite{zhang2020single}. We outperformed MCNL by a large margin,  which gains improvement of 15.4 Rank-1 and 22.3 mAP on Makret-SCT, 13.2 Rank-1 and 19.3 mAP on Duke-SCT, 23.5 Rank-1, 
because our method further exploits cross-camera self supervision information which is ignored in MCNL. 

%Since Precise-ICS\cite{wang2021towards} achieved best performance on ICS-AS, we implemented it in ICS-DS setting. The performance dropped dramatically compared with that in ICS-AS setting. The main reason is the tracklet association methods are not reliable under ICS-DS, they matched two tracklet with different identities into the same identity which resulted in poor performance. In chapter 4.5, we will further show a comparison under an intermediate circumstance between ICS-DS and ICS-AS setting, i.e., partial identities have cross-camera positive pairs while labelled with different categories.

\vspace{0.1cm}\noindent \textbf{Self-supervised learning methods.}
For comparison with self-supervised learning methods, we chose representative contrastive self-supervised learning  method Simsiam \cite{chen2020exploring} and adversarial self-supervised learning method HHL \cite{zhong2018generalizing} that is specific for Re-ID.
For fair comparison, we modified unsupervised HHL as an intra-camera supervised version denoted as $\text{HHL}^{\dagger}$. Specifically, intra-camera positive pairs as well as their generated images shared the same label. We outperformed these two representative self-supervised methods by a large margin. The effectiveness of HHL is restricted by the quality of generated images. Simsiam ignores camera-specific distribution of person Re-ID. In comparison, our cross-camera feature augmentation can model cross-camera relation more effectively.

\begin{table}[ht]
\small
    \centering
    \caption{Ablation study on individual components of CCFP. ``CaCE'' denotes camera-aware cross entropy loss. ``GSL'' denotes global-level self-supervised learning loss.``LSL'' denotes local-level self-supervised learning loss. ``SSRC'' denotes shared-specific region constraint loss.}
    \vspace{-0.2cm}
    \begin{tabular}{c||cc||cc}
         \hline
         \multirow{2}*{Methods}&   \multicolumn{2}{c||}{Market-SCT}   &\multicolumn{2}{c}{Duke-SCT} \\ \cline{2-5}
           &R-1   &mAP  &R-1    &mAP   \\  \hline\hline
        baseline  &75.7 &51.5 &70.1 &53.1\\
        + CaCE &77.6 &52.3&71.4 &54.6 \\
        + CaCE + GSL & 78.9 &59.5 &75.7 &60.4 \\
        %full (w/o L-MCNL)  & 81.2 &61.7 &78.7 &62.9 \\
        + CaCE + GSL + LSL (w/o SSRC)  & 80.5 &61.4 &78.7 &62.7 \\
        + CaCE + GSL + LSL (full model) &\textbf{82.4} &\textbf{63.9} &\textbf{80.3} &\textbf{64.5}\\ \hline

    \end{tabular}

    %Baseline used traditional cross entropy loss (CE) and metric learning loss (i.e., MCNL). 'baseline+CaCE' means we replace traditional CE loss with our proposed CaCE loss. 'baseline+CaCE+GSL' means we add the global-level self-supervised loss upon '+CaCE'. 'baseline+CaCE+GSL+LSL' means we add the local-level self-supervised loss . '+LSL(w/o L-MCNL)' means we delete L-MCNL from LSL. '+LSL(w/o LCR)' means we delete LCR from LSL.}
    \label{ablation}
    \vspace{-0.6cm}   
\end{table}

\vspace{0.1cm}\noindent \textbf{Distribution alignment methods.}
Since our proposed method is closely related with  distribution alignment methods, we also compared two commonly used distribution alignment losses Maximum Mean Discrepancy (MMD) loss \cite{mmd} and CORAL loss \cite{coral}. For fair comparison, both losses were applied based on our baseline. Our method still outperformed them, because our method models the relation between different domains by cross-camera features in instance level. While commonly used distribution alignment loss only considers the global distribution distances.

%In summary, our method outperformed several advanced state-of-the-art methods by a large margin in ICS-DS setting. The main reason is that we mine cross-camera self supervision from camera-specific feature distribution to  make connection between cross-camera unpaired data and facilitate camera-invariant representation learning. 
%In contrast, most of existing methods could not achieve this goal except for GANs-based self-supervised learning methods. However, we also conducted experiments with SimSiam\cite{chen2020exploring} + HHL\cite{zhong2018generalizing} and found it is even inferior to the model trained on ordinary loss by a relative large margin. The main reason is that person Re-ID task aims to retrieve person across camera views with several variations. Augmentations without elaborated design for Re-ID task cannot reflect these variations and even results in wrong guidance.

\subsection{Ablation Study}

We performed ablation study to evaluate the effectiveness of each component in our model. The experimental results are reported in Table \ref{ablation}. 
``+'' means combining the component with baseline.

\begin{figure}[ht]
\includegraphics[width=7.5cm]{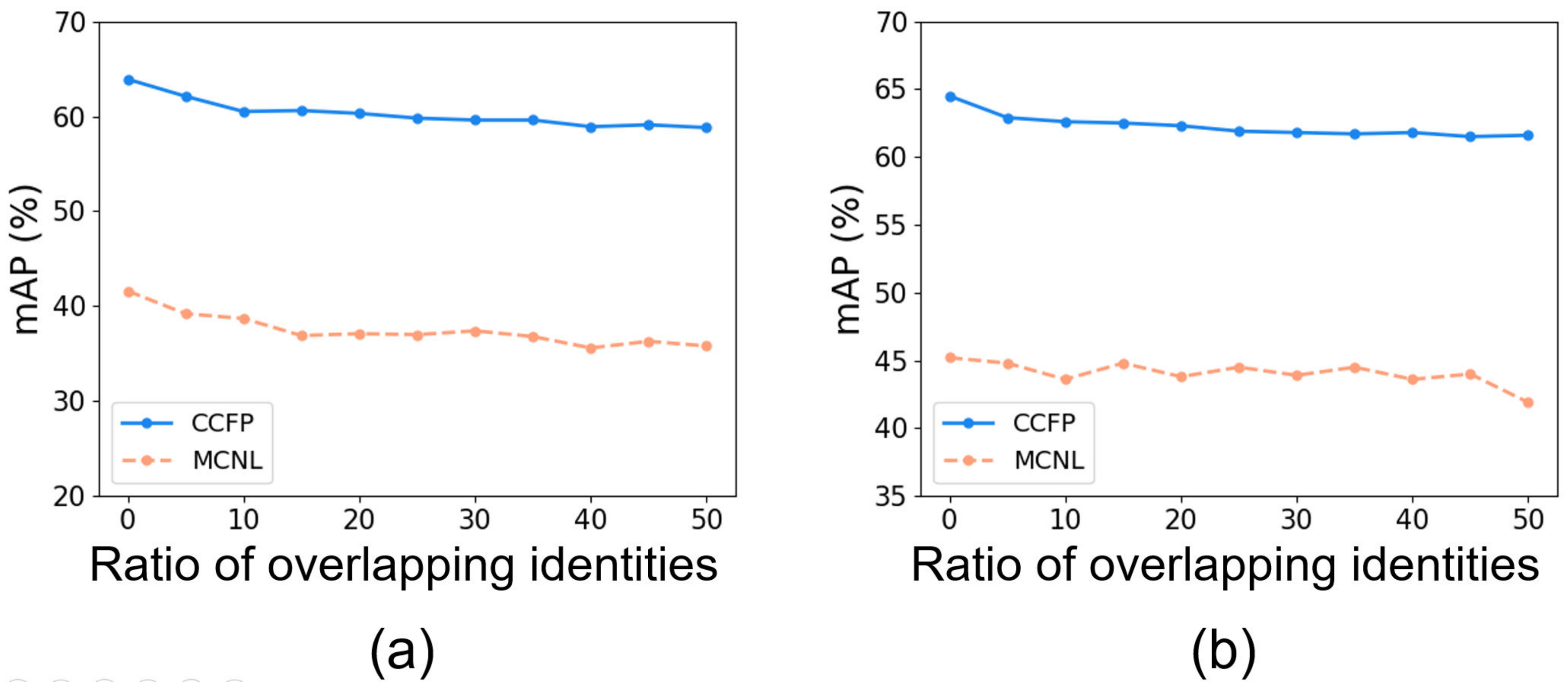}

\vspace{-0.2cm}
\caption{Analysis on ratio of overlapping identities. (a) Comparison between CCFP and MCNL on Market-1501. (b) Comparison between CCFP and MCNL on DukeMTMC.}
\label{overlap}
\vspace{-0.6cm}
\end{figure}

\noindent\textbf{The effectiveness of GSL.}
%We added the 'GSL' upon the 'basic+CaCE' to validate its effectiveness.
Compared with ``+ CaCE'', 
the performance of ``+ CaCE + GSL'' was clearly improved on both datasets. This indicates that global-level self-supervised learning can effectively mine cross-camera self supervision information to eliminate cross-camera scene variations without cross-camera positive pairs. %This is because \modify{ transformed fake cross-camera positive pairs to eliminate cross-camera scene variations and learn camera-invariant representations.}
%GLS can provide cross-camera cues which are crucial for camera-invariant representation learning by mining self supervision information from camera-specific feature distribution which is achieved by generating fake cross-camera positive feature pairs which encode the same identity but in different camera distribution and minimize the distances of the fake pairs. Pulling them together imitates pulling cross-camera pairs together in fully supervised Re-ID.

\vspace{0.1cm}\noindent\textbf{The effectiveness of LSL.}
Compared with ``+ CaCE + GSL'',
the performance of ``+ CaCE + GSL + LSL'' is clearly improved, which indicates that the fine-grained self supervision information  can complement global-level self-supervised learning.
%Complementary to the above-mentioned global-level self-supervised loss, \modify{LSL improved the model by mining more fine-grained self supervision information from automatically localized local regions to facilitate camera-invariant representation learning.}

\vspace{0.1cm}\noindent\textbf{The effectiveness of SSRC.}
Compared with ``+ CaCE + GSL + LSL'', the performance of ``+ CaCE + GSL + LSL (w/o SSRC)'' dropped, which indicates that more diverse local-level feature can facilitate local-level self-supervised learning.
%For multiple local region learning, the SRC loss enforces the selected local-level features to be specific to capture part-specific information, and meanwhile to be shared to avoid attending to person-irrelevant regions for facilitating fine-grained local-level self-supervised learning.} The performance dropped without SRC because the local-level features cannot focus on different regions. In fact, we calculated the SRC without calculating its gradient, the loss was very large which suggested that these local regions were simply focus on the similar regions with less useful information.

%\noindent\textbf{The effectiveness of L-MCNL.}
%On Duke-SCT, the results of "full" beat "full (w/o L-MCNL)" by 1.5 and 1.6 on rank-1 accuracy and mAP, respectively. This demonstrated that investigating fine-grained local-level features is beneficial for metric learning. 

\vspace{0.1cm}\noindent \textbf{The effectiveness of CaCE.}
Compared with ``baseline'' which employed traditional cross entropy loss, the performance of our camera-aware cross entropy ``+ CaCE'' was improved on both datasets, which indicates CaCE is more effective in ICS-DS setting. %complement metric learning loss (i.e., MCNL)  better.

%We trained the model with CaCE instead of traditional cross entropy loss to validate its effectiveness. "basic+CaCE" outperformed the basic model with traditional cross entropy loss. The main reason is that the key idea of the metric learning loss used in our baseline is that enforcing the cross-camera hard negative closer than within-camera hard negative while CaCE assimilated this thought.

\subsection{Further Analysis}

\vspace{0.1cm}\noindent \textbf{The impact of cross-camera identity overlap ratio.} We also consider the situations between ICS-DS setting and ICS-AS setting.
Since in real-world large-scale surveillance systems, matching between scenes of different distances are required, i.e., there are different ratios of overlapping identities across different cameras.
We additionally simulate different ratios of overlapping identities.
%Toward a more practical situation, we reconstructed training set with certain percent overlap rate, i.e., we randomly chose some identities whose images will be cross-camera paired data but with different identities labelled in each camera. We set the overlap rate as $5\% $ to $50\%$ with increasing by $5\%$ each time and carried out experiments on these datasets and show robustness of our model under the outliers.

%In this section, we simulate the real world situation, i.e.,  in large-scale surveillance systems of the real world,matching between adjacent scenes and distant scenes is both required, so some identities may have cross-camera positive pairs while we label them with different classes in each camera view. 
To evaluate our method in this setting, we reconstructed the training set on Market-1501 and DukeMTMC with different ratios of overlapping identities from 5\% to 50\% with increment of $5\%$ in each step. 
The state-of-the-art ICS-DS Re-ID method MCNL was compared.
The results are shown in Figure \ref{overlap}. The results demonstrate that our method was robust to identity overlap ratios, because our method does not rely on  cross-camera overlapping identities.
Moreover, our method outperformed  MCNL by a large margin.
%Since our proposed GSL only enforce positive pairs as well as corresponding cross-camera features close and our LSL only enforce each sample and corresponding cross-camera local-level features close, Both of which do not push negative pairs away.
%As a result, though the samples of the same identity in different cameras are labelled with different classes, we do not push them away. The main reason for a small amount of performance degradation comes from the MCNL in baseline. From the Figure \ref{overlap} we can see, the tendency between CCFP and MCNL is similar.

%relax the hypothesis that each person will only be captured in one camera which is too absolute and carried out experiments on Market-SCT with certain overlap rate which is more practical in the open world. The samples with the same identity in different camera views are labelled with different category. The results are shown in Fig. \ref{overlap}. We point out that our method is robust under the ow-ICS. Since our GSL only enforce positive pairs as well as corresponding cross-camera features close and our LSL only enforce the same sample as well as corresponding cross-camera local region features close. Both of which do not push negative pairs away. As a result, though the samples which share the same identity in different cameras are labelled with different classes, we do not push them away. The main reason for a small amount of performance degradation comes from the CE loss and MCNL in baseline. From the Fig. \ref{overlap} we can see, the tendency between baseline and CCFP is similar.

\begin{figure}[ht]
\includegraphics[width=7.5cm]{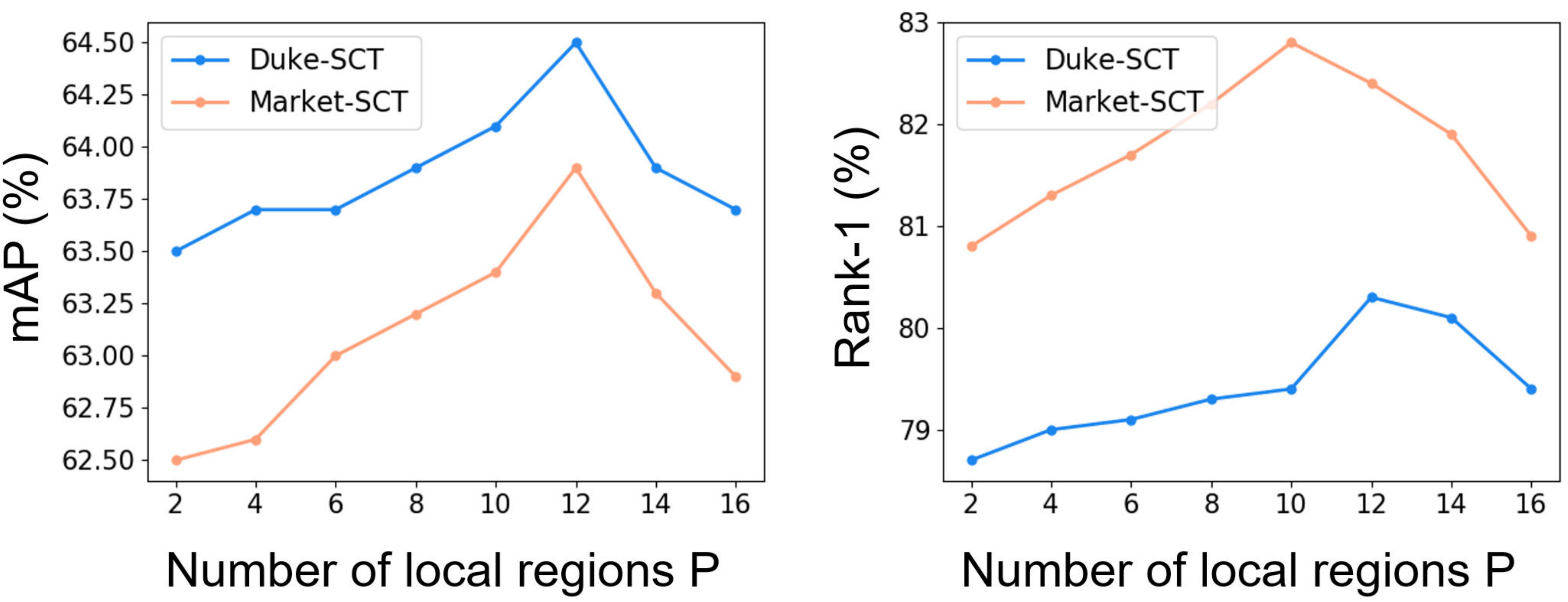}
\vspace{-0.1cm}
\caption{Analysis on the number of local regions $P$.}
\label{patch}
\vspace{-0.3cm}

\end{figure}

\vspace{0.1cm}\noindent \textbf{The impact of number of local regions $P$.}
The number of local regions determines how many local-level features are output by the Transformer.
The results are reported in Figure \ref{patch}. We varied $P$ from 2 to 16.
When $P$ was increased from 2 to 12, the performance kept increasing because more diverse fine-grained local-level features can be extracted for local-level self-supervised learning.
When $P$ was larger than 12, we observed a drop of performance, 
since excessive local regions make local-level features become redundant.
%because some unrelated local-level features were included.

\begin{figure}[ht]

\centering

\includegraphics[width=7.5cm]{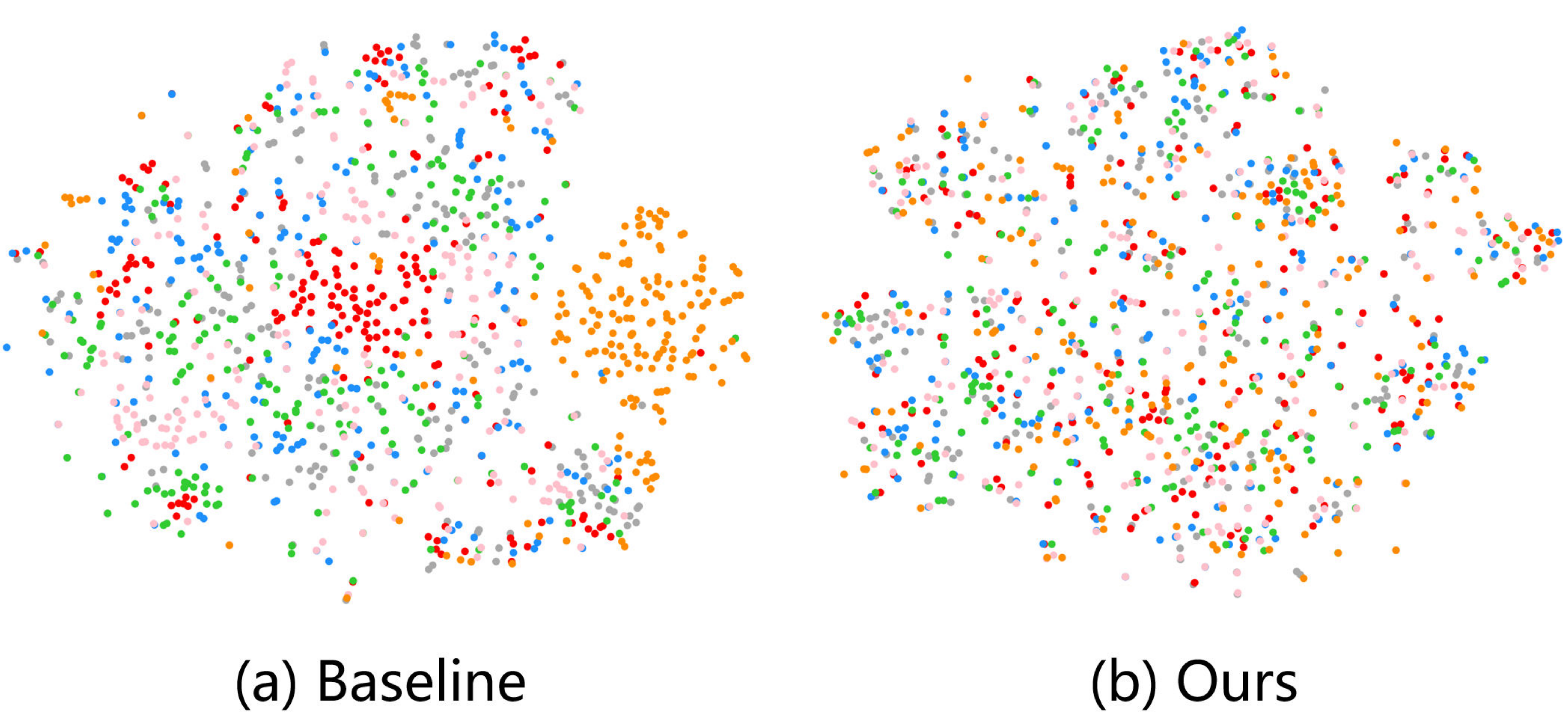}
\vspace{-0.1cm}
\caption{t-SNE visualization \cite{van2008visualizing} of feature distribution from a subset of Market-1501. Best viewed in color.}
\label{distri}
\end{figure}

\vspace{0.1cm}\noindent\textbf{Visualization.}
To better understand the effect of cross-camera feature prediction as for self supervision information mining and learning, we visualized the feature distribution learned by the baseline model and our method as shown in Figure \ref{distri}. From figure \ref{distri} we can observe that, the camera-specific distribution of baseline is not identical, which is harmful for cross-camera matching,  while our method achieves more identical camera-specific distribution, which is favorable for cross-camera matching.

%Figure \ref{distri10} presents the image features of 10 IDs from training set of Market-1501. Compared with the baseline, for our method, the real cross-camera positive pairs (orange) that are missing on Market-SCT are more compact with the intra-camera training samples (blue), which indicates that our transformed fake cross-camera positive features can help learning camera-invariant features as using the missing real cross-camera positive features.

%the baseline separates the samples according to the cameras which is harmful for cross-camera matching. While features from different cameras have similar distribution in our method.

%We also randomly select 10 identities taken from the testing set of Market-1501 to observe the compactness of intra-class and dispersion of inter-class between the baseline and our method. As shown in Figure \ref{10}, our method obviously achieves higher compactness of intra-class and higher dispersion of inter-class.

\section{Conclusion}

In this paper, we investigate intra-camera supervised person re-identification across distant scenes (ICS-DS Re-ID), which is significant for large-scale video surveillance systems but remains under-explored.
Lack of cross-camera paired data leads to a challenge of how to establish the relation of unpaired samples among different cameras, which is ignored by most intra-camera supervised person re-identification methods.
To overcome this issue, we propose a cross-camera feature prediction (CCFP) to mine cross-camera self supervision information from camera-specific distributions by transforming fake cross-camera positive features based on camera-specific batch normalization layers.
Furthermore, to capture more fine-grained cross-camera self supervision information, we employ transformer to automatically localize and extract local-level features for local-level self-supervised learning.
Finally, cross-camera self supervision and intra-camera supervision are aggregated into an unified framework for camera-invariant representation learning. Extensive experiments on Market-SCT, Duke-SCT and MSMT17-SCT in the ICS-DS setting validate the superiority of our method. 

\section*{Acknowledgement}
This work was partially supported by NSFC (U1911401,
U1811461), China National Postdoctoral Program for Innovative Talents (BX2020\\0395), China Postdoctoral Science Foundation (2021M693616),  Guangdong NSF Project (No. 2020B1515120085,
2018B030312002), and the Key-Area Research and Development
Program of Guangzhou (202007030004).
%demonstrate the effectiveness of self-supervised learning via augmentation in the feature-level instead of in the image-level. Specifically, we propose a novel self-supervised learning method with Cross-Camera Features (CCFs) outputted by the Cross-Camera Predictor (CCP). To capture cross-camera cues for SCT person Re-ID, the CCP is composed of $C$ Camera-Specific Batch Normalization (CSBN) layers. Each head estimates the camera-specific statistics for feature normalization to imitate representations under different camera styles. More specifically, we develop a self-supervised learning loss both in global-level and local-level. For the global-level, we insert a CCP in the bottleneck\cite{luo2019bag} after the backbone and enforce the anchor feature outputted by the BNNeck to be pulled close to all the corresponding CCFs. For the local-level, we insert another CCP after the backbone on the feature map level, for each feature map, we input it into the HMP and have $C$ CCFMs. All of them will then input the Transformer to extract local region features which then are forced to get close to the anchor local region features. Extensive experiments validate the effectiveness of our proposed method as well as each component in our model.



\bibliographystyle{ACM-Reference-Format}
\bibliography{ref}


\begin{thebibliography}{64}

%%% the Web address.

\ifx \showCODEN    \undefined \def \showCODEN     #1{\unskip}     \fi
\ifx \showDOI      \undefined \def \showDOI       #1{#1}\fi
\ifx \showISBNx    \undefined \def \showISBNx     #1{\unskip}     \fi
\ifx \showISBNxiii \undefined \def \showISBNxiii  #1{\unskip}     \fi
\ifx \showISSN     \undefined \def \showISSN      #1{\unskip}     \fi
\ifx \showLCCN     \undefined \def \showLCCN      #1{\unskip}     \fi
\ifx \shownote     \undefined \def \shownote      #1{#1}          \fi
\ifx \showarticletitle \undefined \def \showarticletitle #1{#1}   \fi
\ifx \showURL      \undefined \def \showURL       {\relax}        \fi
% invisible to TeX
\providecommand\bibfield[2]{#2}
\providecommand\bibinfo[2]{#2}
\providecommand\natexlab[1]{#1}
\providecommand\showeprint[2][]{arXiv:#2}

\bibitem[\protect\citeauthoryear{Bazzani, Cristani, and Murino}{Bazzani
  et~al\mbox{.}}{2013}]%
        {bazzani2013symmetry}
\bibfield{author}{\bibinfo{person}{Loris Bazzani}, \bibinfo{person}{Marco
  Cristani}, {and} \bibinfo{person}{Vittorio Murino}.}
  \bibinfo{year}{2013}\natexlab{}.
\newblock \showarticletitle{Symmetry-driven accumulation of local features for
  human characterization and re-identification}.
\newblock \bibinfo{journal}{\emph{Computer Vision and Image
  Understanding(CVIU)}} (\bibinfo{year}{2013}).
\newblock


\bibitem[\protect\citeauthoryear{Carion, Massa, Synnaeve, Usunier, Kirillov,
  and Zagoruyko}{Carion et~al\mbox{.}}{2020}]%
        {carion2020end}
\bibfield{author}{\bibinfo{person}{Nicolas Carion}, \bibinfo{person}{Francisco
  Massa}, \bibinfo{person}{Gabriel Synnaeve}, \bibinfo{person}{Nicolas
  Usunier}, \bibinfo{person}{Alexander Kirillov}, {and} \bibinfo{person}{Sergey
  Zagoruyko}.} \bibinfo{year}{2020}\natexlab{}.
\newblock \showarticletitle{End-to-end object detection with transformers}. In
  \bibinfo{booktitle}{\emph{European Conference on Computer Vision (ECCV)}}.
  Springer.
\newblock


\bibitem[\protect\citeauthoryear{Caron, Misra, Mairal, Goyal, Bojanowski, and
  Joulin}{Caron et~al\mbox{.}}{2020}]%
        {caron2020unsupervised}
\bibfield{author}{\bibinfo{person}{Mathilde Caron}, \bibinfo{person}{Ishan
  Misra}, \bibinfo{person}{Julien Mairal}, \bibinfo{person}{Priya Goyal},
  \bibinfo{person}{Piotr Bojanowski}, {and} \bibinfo{person}{Armand Joulin}.}
  \bibinfo{year}{2020}\natexlab{}.
\newblock \showarticletitle{Unsupervised learning of visual features by
  contrasting cluster assignments}.
\newblock \bibinfo{journal}{\emph{arXiv preprint arXiv:2006.09882}}
  (\bibinfo{year}{2020}).
\newblock


\bibitem[\protect\citeauthoryear{Chen, Ding, Xie, Yuan, Chen, Yang, Ren, and
  Wang}{Chen et~al\mbox{.}}{2019}]%
        {chen2019abd}
\bibfield{author}{\bibinfo{person}{Tianlong Chen}, \bibinfo{person}{Shaojin
  Ding}, \bibinfo{person}{Jingyi Xie}, \bibinfo{person}{Ye Yuan},
  \bibinfo{person}{Wuyang Chen}, \bibinfo{person}{Yang Yang},
  \bibinfo{person}{Zhou Ren}, {and} \bibinfo{person}{Zhangyang Wang}.}
  \bibinfo{year}{2019}\natexlab{}.
\newblock \showarticletitle{Abd-net: Attentive but diverse person
  re-identification}. In \bibinfo{booktitle}{\emph{Proceedings of the IEEE
  International Conference on Computer Vision (CVPR)}}.
\newblock


\bibitem[\protect\citeauthoryear{Chen, Kornblith, Norouzi, and Hinton}{Chen
  et~al\mbox{.}}{2020}]%
        {chen2020simple}
\bibfield{author}{\bibinfo{person}{Ting Chen}, \bibinfo{person}{Simon
  Kornblith}, \bibinfo{person}{Mohammad Norouzi}, {and}
  \bibinfo{person}{Geoffrey Hinton}.} \bibinfo{year}{2020}\natexlab{}.
\newblock \showarticletitle{A simple framework for contrastive learning of
  visual representations}. In \bibinfo{booktitle}{\emph{International
  conference on machine learning (ICML)}}. PMLR.
\newblock


\bibitem[\protect\citeauthoryear{Chen and He}{Chen and He}{2020}]%
        {chen2020exploring}
\bibfield{author}{\bibinfo{person}{Xinlei Chen} {and} \bibinfo{person}{Kaiming
  He}.} \bibinfo{year}{2020}\natexlab{}.
\newblock \showarticletitle{Exploring Simple Siamese Representation Learning}.
\newblock \bibinfo{journal}{\emph{arXiv preprint arXiv:2011.10566}}
  (\bibinfo{year}{2020}).
\newblock


\bibitem[\protect\citeauthoryear{{Chen}, {Zhu}, {Zheng}, and {Lai}}{{Chen}
  et~al\mbox{.}}{2018}]%
        {Feature-Augmentation}
\bibfield{author}{\bibinfo{person}{Y. {Chen}}, \bibinfo{person}{X. {Zhu}},
  \bibinfo{person}{W. {Zheng}}, {and} \bibinfo{person}{J. {Lai}}.}
  \bibinfo{year}{2018}\natexlab{}.
\newblock \showarticletitle{Person Re-Identification by Camera Correlation
  Aware Feature Augmentation}.
\newblock \bibinfo{journal}{\emph{IEEE Transactions on Pattern Analysis and
  Machine Intelligence (TPAMI)}} (\bibinfo{year}{2018}).
\newblock


\bibitem[\protect\citeauthoryear{Deng, Dong, Socher, Li, Li, and Fei-Fei}{Deng
  et~al\mbox{.}}{2009}]%
        {deng2009imagenet}
\bibfield{author}{\bibinfo{person}{Jia Deng}, \bibinfo{person}{Wei Dong},
  \bibinfo{person}{Richard Socher}, \bibinfo{person}{Li-Jia Li},
  \bibinfo{person}{Kai Li}, {and} \bibinfo{person}{Li Fei-Fei}.}
  \bibinfo{year}{2009}\natexlab{}.
\newblock \showarticletitle{Imagenet: A large-scale hierarchical image
  database}. In \bibinfo{booktitle}{\emph{IEEE conference on computer vision
  and pattern recognition (CVPR)}}.
\newblock


\bibitem[\protect\citeauthoryear{Deng, Guo, Xue, and Zafeiriou}{Deng
  et~al\mbox{.}}{2019}]%
        {deng2019arcface}
\bibfield{author}{\bibinfo{person}{Jiankang Deng}, \bibinfo{person}{Jia Guo},
  \bibinfo{person}{Niannan Xue}, {and} \bibinfo{person}{Stefanos Zafeiriou}.}
  \bibinfo{year}{2019}\natexlab{}.
\newblock \showarticletitle{Arcface: Additive angular margin loss for deep face
  recognition}. In \bibinfo{booktitle}{\emph{Proceedings of the IEEE Conference
  on Computer Vision and Pattern Recognition (CVPR)}}.
\newblock


\bibitem[\protect\citeauthoryear{Dinh, Krueger, and Bengio}{Dinh
  et~al\mbox{.}}{2014}]%
        {dinh2014nice}
\bibfield{author}{\bibinfo{person}{Laurent Dinh}, \bibinfo{person}{David
  Krueger}, {and} \bibinfo{person}{Yoshua Bengio}.}
  \bibinfo{year}{2014}\natexlab{}.
\newblock \showarticletitle{Nice: Non-linear independent components
  estimation}.
\newblock \bibinfo{journal}{\emph{arXiv preprint arXiv:1410.8516}}
  (\bibinfo{year}{2014}).
\newblock


\bibitem[\protect\citeauthoryear{Dinh, Sohl-Dickstein, and Bengio}{Dinh
  et~al\mbox{.}}{2016}]%
        {dinh2016density}
\bibfield{author}{\bibinfo{person}{Laurent Dinh}, \bibinfo{person}{Jascha
  Sohl-Dickstein}, {and} \bibinfo{person}{Samy Bengio}.}
  \bibinfo{year}{2016}\natexlab{}.
\newblock \showarticletitle{Density estimation using real nvp}.
\newblock \bibinfo{journal}{\emph{arXiv preprint arXiv:1605.08803}}
  (\bibinfo{year}{2016}).
\newblock


\bibitem[\protect\citeauthoryear{Donahue and Simonyan}{Donahue and
  Simonyan}{2019}]%
        {donahue2019large}
\bibfield{author}{\bibinfo{person}{Jeff Donahue} {and} \bibinfo{person}{Karen
  Simonyan}.} \bibinfo{year}{2019}\natexlab{}.
\newblock \showarticletitle{Large scale adversarial representation learning}.
\newblock \bibinfo{journal}{\emph{arXiv preprint arXiv:1907.02544}}
  (\bibinfo{year}{2019}).
\newblock


\bibitem[\protect\citeauthoryear{{Germain}, {Gregor}, {Murray}, and
  {Larochelle}}{{Germain} et~al\mbox{.}}{2015}]%
        {mathieu2015masked}
\bibfield{author}{\bibinfo{person}{Mathieu {Germain}}, \bibinfo{person}{Karol
  {Gregor}}, \bibinfo{person}{Iain {Murray}}, {and} \bibinfo{person}{Hugo
  {Larochelle}}.} \bibinfo{year}{2015}\natexlab{}.
\newblock \showarticletitle{MADE: Masked Autoencoder for Distribution
  Estimation}. In \bibinfo{booktitle}{\emph{Proceedings of The 32nd
  International Conference on Machine Learning(ICML)}}.
\newblock


\bibitem[\protect\citeauthoryear{Gray and Tao}{Gray and Tao}{2008}]%
        {gray2008viewpoint}
\bibfield{author}{\bibinfo{person}{Douglas Gray} {and} \bibinfo{person}{Hai
  Tao}.} \bibinfo{year}{2008}\natexlab{}.
\newblock \showarticletitle{Viewpoint Invariant Pedestrian Recognition with an
  Ensemble of Localized Features}. In \bibinfo{booktitle}{\emph{European
  Conference on Computer Vision (ECCV)}}.
\newblock


\bibitem[\protect\citeauthoryear{Grill, Strub, Altch{\'e}, Tallec, Richemond,
  Buchatskaya, Doersch, Pires, Guo, Azar, et~al\mbox{.}}{Grill
  et~al\mbox{.}}{2020}]%
        {grill2020bootstrap}
\bibfield{author}{\bibinfo{person}{Jean-Bastien Grill},
  \bibinfo{person}{Florian Strub}, \bibinfo{person}{Florent Altch{\'e}},
  \bibinfo{person}{Corentin Tallec}, \bibinfo{person}{Pierre~H Richemond},
  \bibinfo{person}{Elena Buchatskaya}, \bibinfo{person}{Carl Doersch},
  \bibinfo{person}{Bernardo~Avila Pires}, \bibinfo{person}{Zhaohan~Daniel Guo},
  \bibinfo{person}{Mohammad~Gheshlaghi Azar}, {et~al\mbox{.}}}
  \bibinfo{year}{2020}\natexlab{}.
\newblock \showarticletitle{Bootstrap your own latent: A new approach to
  self-supervised learning}.
\newblock \bibinfo{journal}{\emph{arXiv preprint arXiv:2006.07733}}
  (\bibinfo{year}{2020}).
\newblock


\bibitem[\protect\citeauthoryear{He, Fan, Wu, Xie, and Girshick}{He
  et~al\mbox{.}}{2020a}]%
        {he2020momentum}
\bibfield{author}{\bibinfo{person}{Kaiming He}, \bibinfo{person}{Haoqi Fan},
  \bibinfo{person}{Yuxin Wu}, \bibinfo{person}{Saining Xie}, {and}
  \bibinfo{person}{Ross Girshick}.} \bibinfo{year}{2020}\natexlab{a}.
\newblock \showarticletitle{Momentum contrast for unsupervised visual
  representation learning}. In \bibinfo{booktitle}{\emph{Proceedings of the
  IEEE Conference on Computer Vision and Pattern Recognition (CVPR)}}.
\newblock


\bibitem[\protect\citeauthoryear{He, Zhang, Ren, and Sun}{He
  et~al\mbox{.}}{2016}]%
        {he2016deep}
\bibfield{author}{\bibinfo{person}{Kaiming He}, \bibinfo{person}{Xiangyu
  Zhang}, \bibinfo{person}{Shaoqing Ren}, {and} \bibinfo{person}{Jian Sun}.}
  \bibinfo{year}{2016}\natexlab{}.
\newblock \showarticletitle{Deep residual learning for image recognition}. In
  \bibinfo{booktitle}{\emph{Proceedings of the IEEE conference on computer
  vision and pattern recognition (CVPR)}}.
\newblock


\bibitem[\protect\citeauthoryear{He, Liao, Liu, Liu, Cheng, and Mei}{He
  et~al\mbox{.}}{2020b}]%
        {he2020fastreid}
\bibfield{author}{\bibinfo{person}{Lingxiao He}, \bibinfo{person}{Xingyu Liao},
  \bibinfo{person}{Wu Liu}, \bibinfo{person}{Xinchen Liu},
  \bibinfo{person}{Peng Cheng}, {and} \bibinfo{person}{Tao Mei}.}
  \bibinfo{year}{2020}\natexlab{b}.
\newblock \showarticletitle{FastReID: a Pytorch toolbox for real-world person
  re-identification}.
\newblock \bibinfo{journal}{\emph{arXiv preprint arXiv:2006.02631}}
  (\bibinfo{year}{2020}).
\newblock


\bibitem[\protect\citeauthoryear{Ioffe and Szegedy}{Ioffe and Szegedy}{2015}]%
        {ioffe2015batch}
\bibfield{author}{\bibinfo{person}{Sergey Ioffe} {and}
  \bibinfo{person}{Christian Szegedy}.} \bibinfo{year}{2015}\natexlab{}.
\newblock \showarticletitle{Batch normalization: Accelerating deep network
  training by reducing internal covariate shift}. In
  \bibinfo{booktitle}{\emph{International conference on machine learning
  (ICML)}}.
\newblock


\bibitem[\protect\citeauthoryear{Kingma and Ba}{Kingma and Ba}{2014}]%
        {kingma2014adam}
\bibfield{author}{\bibinfo{person}{Diederik~P Kingma} {and}
  \bibinfo{person}{Jimmy Ba}.} \bibinfo{year}{2014}\natexlab{}.
\newblock \showarticletitle{Adam: A method for stochastic optimization}.
\newblock \bibinfo{journal}{\emph{arXiv preprint arXiv:1412.6980}}
  (\bibinfo{year}{2014}).
\newblock


\bibitem[\protect\citeauthoryear{Kingma and Dhariwal}{Kingma and
  Dhariwal}{2018}]%
        {kingma2018glow}
\bibfield{author}{\bibinfo{person}{Diederik~P Kingma} {and}
  \bibinfo{person}{Prafulla Dhariwal}.} \bibinfo{year}{2018}\natexlab{}.
\newblock \showarticletitle{Glow: Generative flow with invertible 1x1
  convolutions}.
\newblock \bibinfo{journal}{\emph{arXiv preprint arXiv:1807.03039}}
  (\bibinfo{year}{2018}).
\newblock


\bibitem[\protect\citeauthoryear{K{\"o}stinger, Hirzer, Wohlhart, Roth, and
  Bischof}{K{\"o}stinger et~al\mbox{.}}{2012}]%
        {2012_CVPR_KISSME}
\bibfield{author}{\bibinfo{person}{Martin K{\"o}stinger},
  \bibinfo{person}{Martin Hirzer}, \bibinfo{person}{Paul Wohlhart},
  \bibinfo{person}{Peter~M Roth}, {and} \bibinfo{person}{Horst Bischof}.}
  \bibinfo{year}{2012}\natexlab{}.
\newblock \showarticletitle{Large scale metric learning from equivalence
  constraints}. In \bibinfo{booktitle}{\emph{IEEE Conference on Computer Vision
  and Pattern Recognition (CVPR)}}.
\newblock


\bibitem[\protect\citeauthoryear{Larsson, Maire, and Shakhnarovich}{Larsson
  et~al\mbox{.}}{2016}]%
        {larsson2016learning}
\bibfield{author}{\bibinfo{person}{Gustav Larsson}, \bibinfo{person}{Michael
  Maire}, {and} \bibinfo{person}{Gregory Shakhnarovich}.}
  \bibinfo{year}{2016}\natexlab{}.
\newblock \showarticletitle{Learning representations for automatic
  colorization}. In \bibinfo{booktitle}{\emph{European conference on computer
  vision (ECCV)}}.
\newblock


\bibitem[\protect\citeauthoryear{Ledig, Theis, Husz{\'a}r, Caballero,
  Cunningham, Acosta, Aitken, Tejani, Totz, Wang, et~al\mbox{.}}{Ledig
  et~al\mbox{.}}{2017}]%
        {ledig2017photo}
\bibfield{author}{\bibinfo{person}{Christian Ledig}, \bibinfo{person}{Lucas
  Theis}, \bibinfo{person}{Ferenc Husz{\'a}r}, \bibinfo{person}{Jose
  Caballero}, \bibinfo{person}{Andrew Cunningham}, \bibinfo{person}{Alejandro
  Acosta}, \bibinfo{person}{Andrew Aitken}, \bibinfo{person}{Alykhan Tejani},
  \bibinfo{person}{Johannes Totz}, \bibinfo{person}{Zehan Wang},
  {et~al\mbox{.}}} \bibinfo{year}{2017}\natexlab{}.
\newblock \showarticletitle{Photo-realistic single image super-resolution using
  a generative adversarial network}. In \bibinfo{booktitle}{\emph{IEEE
  conference on computer vision and pattern recognition (CVPR)}}.
\newblock


\bibitem[\protect\citeauthoryear{{Li}, {Wu}, and {Zheng}}{{Li}
  et~al\mbox{.}}{2021}]%
        {li2021combined}
\bibfield{author}{\bibinfo{person}{Hanjun {Li}}, \bibinfo{person}{Gaojie {Wu}},
  {and} \bibinfo{person}{Wei-Shi {Zheng}}.} \bibinfo{year}{2021}\natexlab{}.
\newblock \showarticletitle{Combined Depth Space based Architecture Search For
  Person Re-identification.}
\newblock \bibinfo{journal}{\emph{arXiv preprint arXiv:2104.04163}}
  (\bibinfo{year}{2021}).
\newblock


\bibitem[\protect\citeauthoryear{Li, Zhu, and Gong}{Li et~al\mbox{.}}{2018}]%
        {tauld}
\bibfield{author}{\bibinfo{person}{Minxian Li}, \bibinfo{person}{Xiatian Zhu},
  {and} \bibinfo{person}{Shaogang Gong}.} \bibinfo{year}{2018}\natexlab{}.
\newblock \showarticletitle{Unsupervised person re-identification by deep
  learning tracklet association}. In \bibinfo{booktitle}{\emph{Proceedings of
  the European conference on computer vision (ECCV)}}.
\newblock


\bibitem[\protect\citeauthoryear{Li, Zhao, Xiao, and Wang}{Li
  et~al\mbox{.}}{2014}]%
        {Li_DeepReID_2014b}
\bibfield{author}{\bibinfo{person}{Wei Li}, \bibinfo{person}{Rui Zhao},
  \bibinfo{person}{Tong Xiao}, {and} \bibinfo{person}{Xiaogang Wang}.}
  \bibinfo{year}{2014}\natexlab{}.
\newblock \showarticletitle{DeepReID: Deep Filter Pairing Neural Network for
  Person Re-identification}. In \bibinfo{booktitle}{\emph{IEEE Conference on
  Computer Vision and Pattern Recognition (CVPR)}}.
\newblock


\bibitem[\protect\citeauthoryear{{Li}, {Zhu}, and {Gong}}{{Li}
  et~al\mbox{.}}{2018}]%
        {li2018harmonious}
\bibfield{author}{\bibinfo{person}{Wei {Li}}, \bibinfo{person}{Xiatian {Zhu}},
  {and} \bibinfo{person}{Shaogang {Gong}}.} \bibinfo{year}{2018}\natexlab{}.
\newblock \showarticletitle{Harmonious Attention Network for Person
  Re-identification}. In \bibinfo{booktitle}{\emph{IEEE Conference on Computer
  Vision and Pattern Recognition (CVPR)}}.
\newblock


\bibitem[\protect\citeauthoryear{Liao, Hu, Zhu, and Li}{Liao
  et~al\mbox{.}}{2015}]%
        {2015_CVPR_LOMO}
\bibfield{author}{\bibinfo{person}{Shengcai Liao}, \bibinfo{person}{Yang Hu},
  \bibinfo{person}{Xiangyu Zhu}, {and} \bibinfo{person}{Stan~Z Li}.}
  \bibinfo{year}{2015}\natexlab{}.
\newblock \showarticletitle{Person re-identification by Local Maximal
  Occurrence representation and metric learning}. In
  \bibinfo{booktitle}{\emph{IEEE Conference on Computer Vision and Pattern
  Recognition (CVPR)}}.
\newblock


\bibitem[\protect\citeauthoryear{Liu, Wen, Yu, Li, Raj, and Song}{Liu
  et~al\mbox{.}}{2017}]%
        {Asoft}
\bibfield{author}{\bibinfo{person}{Weiyang Liu}, \bibinfo{person}{Yandong Wen},
  \bibinfo{person}{Zhiding Yu}, \bibinfo{person}{Ming Li},
  \bibinfo{person}{Bhiksha Raj}, {and} \bibinfo{person}{Le Song}.}
  \bibinfo{year}{2017}\natexlab{}.
\newblock \showarticletitle{Sphereface: Deep hypersphere embedding for face
  recognition}. In \bibinfo{booktitle}{\emph{Proceedings of the IEEE conference
  on computer vision and pattern recognition (CVPR)}}.
\newblock


\bibitem[\protect\citeauthoryear{Liu, Wang, Gong, Lu, and Tao}{Liu
  et~al\mbox{.}}{2019}]%
        {Liu_2019_ICCV}
\bibfield{author}{\bibinfo{person}{Zimo Liu}, \bibinfo{person}{Jingya Wang},
  \bibinfo{person}{Shaogang Gong}, \bibinfo{person}{Huchuan Lu}, {and}
  \bibinfo{person}{Dacheng Tao}.} \bibinfo{year}{2019}\natexlab{}.
\newblock \showarticletitle{Deep Reinforcement Active Learning for
  Human-in-the-Loop Person Re-Identification}. In \bibinfo{booktitle}{\emph{The
  IEEE International Conference on Computer Vision (ICCV)}}.
\newblock


\bibitem[\protect\citeauthoryear{Long, Cao, Wang, and Jordan}{Long
  et~al\mbox{.}}{2015}]%
        {mmd}
\bibfield{author}{\bibinfo{person}{Mingsheng Long}, \bibinfo{person}{Yue Cao},
  \bibinfo{person}{Jianmin Wang}, {and} \bibinfo{person}{Michael Jordan}.}
  \bibinfo{year}{2015}\natexlab{}.
\newblock \showarticletitle{Learning transferable features with deep adaptation
  networks}. In \bibinfo{booktitle}{\emph{International conference on machine
  learning (ICML)}}.
\newblock


\bibitem[\protect\citeauthoryear{Luo, Gu, Liao, Lai, and Jiang}{Luo
  et~al\mbox{.}}{2019}]%
        {luo2019bag}
\bibfield{author}{\bibinfo{person}{Hao Luo}, \bibinfo{person}{Youzhi Gu},
  \bibinfo{person}{Xingyu Liao}, \bibinfo{person}{Shenqi Lai}, {and}
  \bibinfo{person}{Wei Jiang}.} \bibinfo{year}{2019}\natexlab{}.
\newblock \showarticletitle{Bag of tricks and a strong baseline for deep person
  re-identification}. In \bibinfo{booktitle}{\emph{Proceedings of the IEEE
  Conference on Computer Vision and Pattern Recognition Workshops (CVPR)}}.
\newblock


\bibitem[\protect\citeauthoryear{Luo, Stenger, Zhao, and Kim}{Luo
  et~al\mbox{.}}{2018}]%
        {luo2018trajectories}
\bibfield{author}{\bibinfo{person}{Wenhan Luo}, \bibinfo{person}{Bj{\"o}rn
  Stenger}, \bibinfo{person}{Xiaowei Zhao}, {and} \bibinfo{person}{Tae-Kyun
  Kim}.} \bibinfo{year}{2018}\natexlab{}.
\newblock \showarticletitle{Trajectories as topics: Multi-object tracking by
  topic discovery}.
\newblock \bibinfo{journal}{\emph{IEEE Transactions on Image Processing (TIP)}}
  (\bibinfo{year}{2018}).
\newblock


\bibitem[\protect\citeauthoryear{Qi, Wang, Huo, Shi, and Gao}{Qi
  et~al\mbox{.}}{2019}]%
        {qi2019adversarial}
\bibfield{author}{\bibinfo{person}{Lei Qi}, \bibinfo{person}{Lei Wang},
  \bibinfo{person}{Jing Huo}, \bibinfo{person}{Yinghuan Shi}, {and}
  \bibinfo{person}{Yang Gao}.} \bibinfo{year}{2019}\natexlab{}.
\newblock \showarticletitle{Adversarial camera alignment network for
  unsupervised cross-camera person re-identification}.
\newblock \bibinfo{journal}{\emph{arXiv preprint arXiv:1908.00862}}
  (\bibinfo{year}{2019}).
\newblock


\bibitem[\protect\citeauthoryear{Qi, Wang, Huo, Shi, and Gao}{Qi
  et~al\mbox{.}}{2020}]%
        {qi2020progressive}
\bibfield{author}{\bibinfo{person}{Lei Qi}, \bibinfo{person}{Lei Wang},
  \bibinfo{person}{Jing Huo}, \bibinfo{person}{Yinghuan Shi}, {and}
  \bibinfo{person}{Yang Gao}.} \bibinfo{year}{2020}\natexlab{}.
\newblock \showarticletitle{Progressive cross-camera soft-label learning for
  semi-supervised person re-identification}.
\newblock \bibinfo{journal}{\emph{IEEE Transactions on Circuits and Systems for
  Video Technology (TCSVT)}} (\bibinfo{year}{2020}).
\newblock


\bibitem[\protect\citeauthoryear{Razavi, Oord, and Vinyals}{Razavi
  et~al\mbox{.}}{2019}]%
        {razavi2019generating}
\bibfield{author}{\bibinfo{person}{Ali Razavi}, \bibinfo{person}{Aaron van~den
  Oord}, {and} \bibinfo{person}{Oriol Vinyals}.}
  \bibinfo{year}{2019}\natexlab{}.
\newblock \showarticletitle{Generating diverse high-fidelity images with
  vq-vae-2}.
\newblock \bibinfo{journal}{\emph{arXiv preprint arXiv:1906.00446}}
  (\bibinfo{year}{2019}).
\newblock


\bibitem[\protect\citeauthoryear{{Subramaniam}, {Chatterjee}, and
  {Mittal}}{{Subramaniam} et~al\mbox{.}}{2016}]%
        {subramaniam2016deep}
\bibfield{author}{\bibinfo{person}{Arulkumar {Subramaniam}},
  \bibinfo{person}{Moitreya {Chatterjee}}, {and} \bibinfo{person}{Anurag
  {Mittal}}.} \bibinfo{year}{2016}\natexlab{}.
\newblock \showarticletitle{Deep Neural Networks with Inexact Matching for
  Person Re-Identification}. In \bibinfo{booktitle}{\emph{Advances in Neural
  Information Processing Systems (NeurIPS)}}.
\newblock


\bibitem[\protect\citeauthoryear{Suh, Wang, Tang, Mei, and Lee}{Suh
  et~al\mbox{.}}{2018}]%
        {suh2018part}
\bibfield{author}{\bibinfo{person}{Yumin Suh}, \bibinfo{person}{Jingdong Wang},
  \bibinfo{person}{Siyu Tang}, \bibinfo{person}{Tao Mei}, {and}
  \bibinfo{person}{Kyoung~Mu Lee}.} \bibinfo{year}{2018}\natexlab{}.
\newblock \showarticletitle{Part-aligned bilinear representations for person
  re-identification}. In \bibinfo{booktitle}{\emph{Proceedings of the European
  Conference on Computer Vision (ECCV)}}.
\newblock


\bibitem[\protect\citeauthoryear{Sun and Saenko}{Sun and Saenko}{2016}]%
        {coral}
\bibfield{author}{\bibinfo{person}{Baochen Sun} {and} \bibinfo{person}{Kate
  Saenko}.} \bibinfo{year}{2016}\natexlab{}.
\newblock \showarticletitle{Deep coral: Correlation alignment for deep domain
  adaptation}. In \bibinfo{booktitle}{\emph{European conference on computer
  vision (ECCV)}}.
\newblock


\bibitem[\protect\citeauthoryear{Sun, Zheng, Yang, Tian, and Wang}{Sun
  et~al\mbox{.}}{2018}]%
        {sun2018beyond}
\bibfield{author}{\bibinfo{person}{Yifan Sun}, \bibinfo{person}{Liang Zheng},
  \bibinfo{person}{Yi Yang}, \bibinfo{person}{Qi Tian}, {and}
  \bibinfo{person}{Shengjin Wang}.} \bibinfo{year}{2018}\natexlab{}.
\newblock \showarticletitle{Beyond part models: Person retrieval with refined
  part pooling (and a strong convolutional baseline)}. In
  \bibinfo{booktitle}{\emph{Proceedings of the European conference on computer
  vision (ECCV)}}.
\newblock


\bibitem[\protect\citeauthoryear{Tian, Sun, Poole, Krishnan, Schmid, and
  Isola}{Tian et~al\mbox{.}}{2020}]%
        {tian2020makes}
\bibfield{author}{\bibinfo{person}{Yonglong Tian}, \bibinfo{person}{Chen Sun},
  \bibinfo{person}{Ben Poole}, \bibinfo{person}{Dilip Krishnan},
  \bibinfo{person}{Cordelia Schmid}, {and} \bibinfo{person}{Phillip Isola}.}
  \bibinfo{year}{2020}\natexlab{}.
\newblock \showarticletitle{What makes for good views for contrastive
  learning}.
\newblock \bibinfo{journal}{\emph{arXiv preprint arXiv:2005.10243}}
  (\bibinfo{year}{2020}).
\newblock


\bibitem[\protect\citeauthoryear{Van~der Maaten and Hinton}{Van~der Maaten and
  Hinton}{2008}]%
        {van2008visualizing}
\bibfield{author}{\bibinfo{person}{Laurens Van~der Maaten} {and}
  \bibinfo{person}{Geoffrey Hinton}.} \bibinfo{year}{2008}\natexlab{}.
\newblock \showarticletitle{Visualizing data using t-SNE.}
\newblock \bibinfo{journal}{\emph{Journal of machine learning research (JMLR)}}
  (\bibinfo{year}{2008}).
\newblock


\bibitem[\protect\citeauthoryear{Vaswani, Shazeer, Parmar, Uszkoreit, Jones,
  Gomez, Kaiser, and Polosukhin}{Vaswani et~al\mbox{.}}{2017}]%
        {vaswani2017attention}
\bibfield{author}{\bibinfo{person}{Ashish Vaswani}, \bibinfo{person}{Noam
  Shazeer}, \bibinfo{person}{Niki Parmar}, \bibinfo{person}{Jakob Uszkoreit},
  \bibinfo{person}{Llion Jones}, \bibinfo{person}{Aidan~N Gomez},
  \bibinfo{person}{Lukasz Kaiser}, {and} \bibinfo{person}{Illia Polosukhin}.}
  \bibinfo{year}{2017}\natexlab{}.
\newblock \showarticletitle{Attention is all you need}.
\newblock \bibinfo{journal}{\emph{arXiv preprint arXiv:1706.03762}}
  (\bibinfo{year}{2017}).
\newblock


\bibitem[\protect\citeauthoryear{Wang, Yuan, Chen, Li, and Zhou}{Wang
  et~al\mbox{.}}{2018}]%
        {MGN}
\bibfield{author}{\bibinfo{person}{Guanshuo Wang}, \bibinfo{person}{Yufeng
  Yuan}, \bibinfo{person}{Xiong Chen}, \bibinfo{person}{Jiwei Li}, {and}
  \bibinfo{person}{Xi Zhou}.} \bibinfo{year}{2018}\natexlab{}.
\newblock \showarticletitle{Learning discriminative features with multiple
  granularities for person re-identification}. In
  \bibinfo{booktitle}{\emph{Proceedings of the 26th ACM international
  conference on Multimedia (ACMMM)}}.
\newblock


\bibitem[\protect\citeauthoryear{Wang, Lai, Chen, Huang, Gong, and Hua}{Wang
  et~al\mbox{.}}{2021}]%
        {wang2021towards}
\bibfield{author}{\bibinfo{person}{Menglin Wang}, \bibinfo{person}{Baisheng
  Lai}, \bibinfo{person}{Haokun Chen}, \bibinfo{person}{Jianqiang Huang},
  \bibinfo{person}{Xiaojin Gong}, {and} \bibinfo{person}{Xian-Sheng Hua}.}
  \bibinfo{year}{2021}\natexlab{}.
\newblock \showarticletitle{Towards precise intra-camera supervised person
  re-identification}. In \bibinfo{booktitle}{\emph{Proceedings of the IEEE
  Winter Conference on Applications of Computer Vision (WACV)}}.
\newblock


\bibitem[\protect\citeauthoryear{{Wang}, {Gong}, {Zhu}, and {Wang}}{{Wang}
  et~al\mbox{.}}{2014}]%
        {wang2014person}
\bibfield{author}{\bibinfo{person}{Taiqing {Wang}}, \bibinfo{person}{Shaogang
  {Gong}}, \bibinfo{person}{Xiatian {Zhu}}, {and} \bibinfo{person}{Shengjin
  {Wang}}.} \bibinfo{year}{2014}\natexlab{}.
\newblock \showarticletitle{Person re-identification by video ranking}. In
  \bibinfo{booktitle}{\emph{European Conference on Computer Vision (ECCV)}}.
\newblock


\bibitem[\protect\citeauthoryear{{Wang}, {Gong}, {Zhu}, and {Wang}}{{Wang}
  et~al\mbox{.}}{2016}]%
        {wang2016person}
\bibfield{author}{\bibinfo{person}{Taiqing {Wang}}, \bibinfo{person}{Shaogang
  {Gong}}, \bibinfo{person}{Xiatian {Zhu}}, {and} \bibinfo{person}{Shengjin
  {Wang}}.} \bibinfo{year}{2016}\natexlab{}.
\newblock \showarticletitle{Person Re-Identification by Discriminative
  Selection in Video Ranking}.
\newblock \bibinfo{journal}{\emph{IEEE Transactions on Pattern Analysis and
  Machine Intelligence (TPAMI)}} (\bibinfo{year}{2016}).
\newblock


\bibitem[\protect\citeauthoryear{Wei, Zhang, Gao, and Tian}{Wei
  et~al\mbox{.}}{2018}]%
        {wei2018person}
\bibfield{author}{\bibinfo{person}{Longhui Wei}, \bibinfo{person}{Shiliang
  Zhang}, \bibinfo{person}{Wen Gao}, {and} \bibinfo{person}{Qi Tian}.}
  \bibinfo{year}{2018}\natexlab{}.
\newblock \showarticletitle{Person transfer gan to bridge domain gap for person
  re-identification}. In \bibinfo{booktitle}{\emph{IEEE Conference on Computer
  Vision and Pattern Recognition (CVPR)}}.
\newblock


\bibitem[\protect\citeauthoryear{Wen, Zhang, Li, and Qiao}{Wen
  et~al\mbox{.}}{2016}]%
        {center}
\bibfield{author}{\bibinfo{person}{Yandong Wen}, \bibinfo{person}{Kaipeng
  Zhang}, \bibinfo{person}{Zhifeng Li}, {and} \bibinfo{person}{Yu Qiao}.}
  \bibinfo{year}{2016}\natexlab{}.
\newblock \showarticletitle{A discriminative feature learning approach for deep
  face recognition}. In \bibinfo{booktitle}{\emph{European conference on
  computer vision (ECCV)}}.
\newblock


\bibitem[\protect\citeauthoryear{Wu, Zhu, and Gong}{Wu et~al\mbox{.}}{2020}]%
        {tssl}
\bibfield{author}{\bibinfo{person}{Guile Wu}, \bibinfo{person}{Xiatian Zhu},
  {and} \bibinfo{person}{Shaogang Gong}.} \bibinfo{year}{2020}\natexlab{}.
\newblock \showarticletitle{Tracklet self-supervised learning for unsupervised
  person re-identification}. In \bibinfo{booktitle}{\emph{Proceedings of the
  AAAI Conference on Artificial Intelligence (AAAI)}}.
\newblock


\bibitem[\protect\citeauthoryear{Wu, Yang, Liu, Liao, Lei, and Li}{Wu
  et~al\mbox{.}}{2019}]%
        {graph_asso}
\bibfield{author}{\bibinfo{person}{Jinlin Wu}, \bibinfo{person}{Yang Yang},
  \bibinfo{person}{Hao Liu}, \bibinfo{person}{Shengcai Liao},
  \bibinfo{person}{Zhen Lei}, {and} \bibinfo{person}{Stan~Z Li}.}
  \bibinfo{year}{2019}\natexlab{}.
\newblock \showarticletitle{Unsupervised graph association for person
  re-identification}. In \bibinfo{booktitle}{\emph{IEEE International
  Conference on Computer Vision (ICCV)}}.
\newblock


\bibitem[\protect\citeauthoryear{Xiao, Li, Ouyang, and Wang}{Xiao
  et~al\mbox{.}}{2016}]%
        {2016_CVPR_JSTL}
\bibfield{author}{\bibinfo{person}{Tong Xiao}, \bibinfo{person}{Hongsheng Li},
  \bibinfo{person}{Wanli Ouyang}, {and} \bibinfo{person}{Xiaogang Wang}.}
  \bibinfo{year}{2016}\natexlab{}.
\newblock \showarticletitle{Learning Deep Feature Representations with Domain
  Guided Dropout for Person Re-identification}. In
  \bibinfo{booktitle}{\emph{IEEE Conference on Computer Vision and Pattern
  Recognition (CVPR)}}.
\newblock


\bibitem[\protect\citeauthoryear{Xiong, Gou, Camps, and Sznaier}{Xiong
  et~al\mbox{.}}{2014}]%
        {xiong2014person}
\bibfield{author}{\bibinfo{person}{Fei Xiong}, \bibinfo{person}{Mengran Gou},
  \bibinfo{person}{Octavia Camps}, {and} \bibinfo{person}{Mario Sznaier}.}
  \bibinfo{year}{2014}\natexlab{}.
\newblock \showarticletitle{Person re-identification using kernel-based metric
  learning methods}. In \bibinfo{booktitle}{\emph{European Conference on
  Computer Vision (ECCV)}}.
\newblock


\bibitem[\protect\citeauthoryear{Ye, Shen, Lin, Xiang, Shao, and Hoi}{Ye
  et~al\mbox{.}}{2021}]%
        {ye2021deep}
\bibfield{author}{\bibinfo{person}{Mang Ye}, \bibinfo{person}{Jianbing Shen},
  \bibinfo{person}{Gaojie Lin}, \bibinfo{person}{Tao Xiang},
  \bibinfo{person}{Ling Shao}, {and} \bibinfo{person}{Steven~CH Hoi}.}
  \bibinfo{year}{2021}\natexlab{}.
\newblock \showarticletitle{Deep learning for person re-identification: A
  survey and outlook}.
\newblock \bibinfo{journal}{\emph{IEEE Transactions on Pattern Analysis and
  Machine Intelligence (TPAMI)}} (\bibinfo{year}{2021}).
\newblock


\bibitem[\protect\citeauthoryear{Zhang, Xie, Wei, Zhang, Li, and Tian}{Zhang
  et~al\mbox{.}}{2020}]%
        {zhang2020single}
\bibfield{author}{\bibinfo{person}{Tianyu Zhang}, \bibinfo{person}{Lingxi Xie},
  \bibinfo{person}{Longhui Wei}, \bibinfo{person}{Yongfei Zhang},
  \bibinfo{person}{Bo Li}, {and} \bibinfo{person}{Qi Tian}.}
  \bibinfo{year}{2020}\natexlab{}.
\newblock \showarticletitle{Single camera training for person
  re-identification}. In \bibinfo{booktitle}{\emph{Proceedings of the AAAI
  Conference on Artificial Intelligence (AAAI)}}.
\newblock


\bibitem[\protect\citeauthoryear{{Zheng}, {Shen}, {Tian}, {Wang}, {Wang}, and
  {Tian}}{{Zheng} et~al\mbox{.}}{2015}]%
        {zheng2015scalable}
\bibfield{author}{\bibinfo{person}{Liang {Zheng}}, \bibinfo{person}{Liyue
  {Shen}}, \bibinfo{person}{Lu {Tian}}, \bibinfo{person}{Shengjin {Wang}},
  \bibinfo{person}{Jingdong {Wang}}, {and} \bibinfo{person}{Qi {Tian}}.}
  \bibinfo{year}{2015}\natexlab{}.
\newblock \showarticletitle{Scalable Person Re-identification: A Benchmark}. In
  \bibinfo{booktitle}{\emph{IEEE International Conference on Computer Vision
  (ICCV)}}.
\newblock


\bibitem[\protect\citeauthoryear{Zheng, Shen, Tian, Wang, Wang, and Tian}{Zheng
  et~al\mbox{.}}{2015}]%
        {2015_ICCV_MARKET}
\bibfield{author}{\bibinfo{person}{Liang Zheng}, \bibinfo{person}{Liyue Shen},
  \bibinfo{person}{Lu Tian}, \bibinfo{person}{Shengjin Wang},
  \bibinfo{person}{Jingdong Wang}, {and} \bibinfo{person}{Qi Tian}.}
  \bibinfo{year}{2015}\natexlab{}.
\newblock \showarticletitle{Scalable person re-identification: A benchmark}. In
  \bibinfo{booktitle}{\emph{IEEE International Conference on Computer Vision
  (ICCV)}}.
\newblock


\bibitem[\protect\citeauthoryear{Zheng, Gong, and Xiang}{Zheng
  et~al\mbox{.}}{2013}]%
        {zheng2013RDC}
\bibfield{author}{\bibinfo{person}{Wei-Shi Zheng}, \bibinfo{person}{Shaogang
  Gong}, {and} \bibinfo{person}{Tao Xiang}.} \bibinfo{year}{2013}\natexlab{}.
\newblock \showarticletitle{Reidentification by Relative Distance Comparison}.
\newblock \bibinfo{journal}{\emph{IEEE Transactions on Pattern Analysis and
  Machine Intelligence (TPAMI)}} (\bibinfo{year}{2013}).
\newblock


\bibitem[\protect\citeauthoryear{{Zheng}, {Zheng}, and {Yang}}{{Zheng}
  et~al\mbox{.}}{2017}]%
        {duke}
\bibfield{author}{\bibinfo{person}{Zhedong {Zheng}}, \bibinfo{person}{Liang
  {Zheng}}, {and} \bibinfo{person}{Yi {Yang}}.}
  \bibinfo{year}{2017}\natexlab{}.
\newblock \showarticletitle{Unlabeled Samples Generated by GAN Improve the
  Person Re-identification Baseline in Vitro}. In
  \bibinfo{booktitle}{\emph{2017 IEEE International Conference on Computer
  Vision (ICCV)}}.
\newblock


\bibitem[\protect\citeauthoryear{Zhong, Zheng, Li, and Yang}{Zhong
  et~al\mbox{.}}{2018}]%
        {zhong2018generalizing}
\bibfield{author}{\bibinfo{person}{Zhun Zhong}, \bibinfo{person}{Liang Zheng},
  \bibinfo{person}{Shaozi Li}, {and} \bibinfo{person}{Yi Yang}.}
  \bibinfo{year}{2018}\natexlab{}.
\newblock \showarticletitle{Generalizing a person retrieval model hetero-and
  homogeneously}. In \bibinfo{booktitle}{\emph{Proceedings of the European
  Conference on Computer Vision (ECCV)}}.
\newblock


\bibitem[\protect\citeauthoryear{Zhu, Zhu, Li, Murino, and Gong}{Zhu
  et~al\mbox{.}}{2019}]%
        {zhu2019intra}
\bibfield{author}{\bibinfo{person}{Xiangping Zhu}, \bibinfo{person}{Xiatian
  Zhu}, \bibinfo{person}{Minxian Li}, \bibinfo{person}{Vittorio Murino}, {and}
  \bibinfo{person}{Shaogang Gong}.} \bibinfo{year}{2019}\natexlab{}.
\newblock \showarticletitle{Intra-camera supervised person re-identification: A
  new benchmark}. In \bibinfo{booktitle}{\emph{IEEE International Conference on
  Computer Vision Workshops (ICCV)}}.
\newblock


\bibitem[\protect\citeauthoryear{{Zhuang}, {Wei}, {Xie}, {Ai}, and
  {Tian}}{{Zhuang} et~al\mbox{.}}{2021}]%
        {zhuang2021camera}
\bibfield{author}{\bibinfo{person}{Zijie {Zhuang}}, \bibinfo{person}{Longhui
  {Wei}}, \bibinfo{person}{Lingxi {Xie}}, \bibinfo{person}{Haizhou {Ai}}, {and}
  \bibinfo{person}{Qi {Tian}}.} \bibinfo{year}{2021}\natexlab{}.
\newblock \showarticletitle{Camera-based Batch Normalization: An Effective
  Distribution Alignment Method for Person Re-identification}.
\newblock \bibinfo{journal}{\emph{IEEE Transactions on Circuits and Systems for
  Video Technology(TCSVT)}} (\bibinfo{year}{2021}).
\newblock


\bibitem[\protect\citeauthoryear{Zhuang, Wei, Xie, Zhang, Zhang, Wu, Ai, and
  Tian}{Zhuang et~al\mbox{.}}{2020}]%
        {zhuang2020rethinking}
\bibfield{author}{\bibinfo{person}{Zijie Zhuang}, \bibinfo{person}{Longhui
  Wei}, \bibinfo{person}{Lingxi Xie}, \bibinfo{person}{Tianyu Zhang},
  \bibinfo{person}{Hengheng Zhang}, \bibinfo{person}{Haozhe Wu},
  \bibinfo{person}{Haizhou Ai}, {and} \bibinfo{person}{Qi Tian}.}
  \bibinfo{year}{2020}\natexlab{}.
\newblock \showarticletitle{Rethinking the distribution gap of person
  re-identification with camera-based batch normalization}. In
  \bibinfo{booktitle}{\emph{European Conference on Computer Vision (ECCV)}}.
\newblock


\end{thebibliography}

\end{document}